\documentclass{article} 
\usepackage{aaai25}  
\usepackage{times}  
\usepackage{helvet}  
\usepackage{courier}  
\usepackage[hyphens]{url} 
\usepackage{graphicx} 
\usepackage{natbib}  
\usepackage{caption} 
\usepackage[utf8]{inputenc}
\usepackage{algorithm}
\usepackage[noend]{algpseudocode}
\usepackage{dsfont}
\usepackage{float}
\usepackage{makecell}

\usepackage[switch]{lineno}

\usepackage{amsmath, amsfonts}
\usepackage{booktabs}

\usepackage{xcolor}

\usepackage{enumitem}
\usepackage{multirow}
\usepackage{colortbl}
\usepackage{microtype}
\usepackage{comment}
\usepackage[hang,flushmargin]{footmisc}

\usepackage{amsfonts}

\setlist[itemize]{align=parleft,left=0pt,topsep=1mm,itemsep=0mm,parsep=1mm}

\definecolor{azure(colorwheel)}{rgb}{0.0, 0.5, 1.0}
\definecolor{R5}{rgb}{0.0, 0.7, 0.1}
\definecolor{yw}{rgb}{0.01176, 0.5490, 0.5490}
\definecolor{R123}{rgb}{0.36, 0.54, 0.66}
\definecolor{R1234}{rgb}{0.7, 0.75, 0.71}
\definecolor{applegreen}{rgb}{0.55, 0.71, 0.0}
\definecolor{R132}{rgb}{0.0, 0.0, 1.0}
\definecolor{postechred}{rgb}{0.784, 0.003, 0.313}
\definecolor{gu}{rgb}{0.5460, 0.1755, 0.2766}
\definecolor{el}{rgb}{0.9764, 0.447, 0.447}
\definecolor{hyos}{rgb}{0.662, 0.482, 0.960}
\definecolor{ballblue}{rgb}{0.13, 0.67, 0.8}
\definecolor{cornellred}{rgb}{0.7, 0.11, 0.11}
\definecolor{darkcyan}{rgb}{0.0, 0.55, 0.55}
\definecolor{CuGray}{gray}{0.9}
\definecolor{airforceblue}{rgb}{0.36, 0.54, 0.66}
\definecolor{rev}{rgb}{0.784, 0.003, 0.313}
\definecolor{pink}{cmyk}{0, 0.7808, 0.4429, 0.1412}
\definecolor{amethyst}{rgb}{0.6, 0.4, 0.8}
\definecolor{black}{rgb}{0.0, 0.0, 0.0}
\definecolor{tb3_yellow}{rgb}{0.996, 1.0, 0.6}
\definecolor{R123}{rgb}{0.980, 0.8, 0.604}
\definecolor{R512}{rgb}{0.972, 0.6, 0.6}
\definecolor{dimgray}{rgb}{0.41, 0.41, 0.41}
\definecolor{R3}{rgb}{0.8, 0.25, 0.33}
\definecolor{bleudefrance}{rgb}{0.19, 0.55, 0.91}
\definecolor{R6}{rgb}{0.265, 0.445, 0.765}
\definecolor{blue(ryb)}{rgb}{0.01, 0.28, 1.0}
\definecolor{R4}{rgb}{1.0, 0.49, 0.0}
\definecolor{Gray}{gray}{0.88}
\definecolor{green(ncs)}{rgb}{0.0, 0.62, 0.42}
\definecolor{brightpink}{rgb}{1.0, 0.0, 0.5}
\definecolor{alizarin}{rgb}{0.82, 0.1, 0.26}
\definecolor{coral}{rgb}{1.0,0.412,0.380}

\definecolor{kellygreen}{rgb}{0.3, 0.73, 0.09}

\newcolumntype{g}{>{\columncolor{CuGray}}c}
\newcolumntype{z}{>{\columncolor{CuGray}}l}

\renewcommand{\paragraph}[1]{\vspace{1mm}\noindent\textbf{#1.}\,}

\newcommand{\coral}[1]{\textcolor{coral}{#1}}

\usepackage{xspace}

\makeatletter
\def\@fnsymbol#1{\ensuremath{\ifcase#1\or *\or \dagger\or \ddagger\or
   \mathsection\or \mathparagraph\or \|\or **\or \dagger\dagger
   \or \ddagger\ddagger \else\@ctrerr\fi}}
\makeatother

\def\onedot{.\@\xspace}
\def\eg{\emph{e.g}\onedot} 
\def\ie{\emph{i.e}\onedot}

\newcommand{\Sref}[1]{Sec.~\ref{#1}}
\newcommand{\Eref}[1]{Eq.~(\ref{#1})}
\newcommand{\Fref}[1]{Fig.~\ref{#1}}
\newcommand{\Tref}[1]{Table~\ref{#1}}

\newcommand{\be}{\begin{eqnarray}}
\newcommand{\ee}{\end{eqnarray}}
\newcommand{\bee}{\begin{eqnarray*}}
\newcommand{\eee}{\end{eqnarray*}}

\newcommand{\matrixb}{\left[ \begin{array}}
\newcommand{\matrixe}{\end{array} \right]}

\newcommand{\argmax}{\operatornamewithlimits{\arg \max}}
\newcommand{\argmin}{\operatornamewithlimits{\arg \min}}

\newcommand{\para}[1]{\paragraph{#1}}

\usepackage{amssymb}
\usepackage{pifont}
\newcommand{\cmark}{\ding{51}}%
\newcommand{\xmark}{\ding{55}}%

\definecolor{cvprblue}{rgb}{0.21,0.49,0.74}
\usepackage[pagebackref,breaklinks,colorlinks,citecolor=cvprblue]{hyperref}
\hypersetup{linkcolor=black} %

\usepackage{newfloat}
\usepackage{listings}
\DeclareCaptionStyle{ruled}{labelfont=normalfont,labelsep=colon,strut=off} 
\floatstyle{ruled}
\newfloat{listing}{tb}{lst}{}
\floatname{listing}{Listing}

\title{Zero-shot Depth Completion via Test-time Alignment\\ with Affine-invariant Depth Prior}
\author{
    Lee Hyoseok\textsuperscript{\rm 1},
    Kyeong Seon Kim\textsuperscript{\rm 2},
    Kwon Byung-Ki\textsuperscript{\rm 1},
    Tae-Hyun Oh\textsuperscript{\rm 1,2,3}
}
\affiliations{
    \textsuperscript{\rm 1}Grad.School of Artificial Intelligence  and \textsuperscript{\rm 2}Dept. of Electrical Engineering, POSTECH\\
    \textsuperscript{\rm 3}Institute for Convergence Research and Education in Advanced Technology, Yonsei University\\
    hyos99@postech.ac.kr, ella94.ai@gmail.com, byungki.kwon@postech.ac.kr,
    taehyun@postech.ac.kr
}

\usepackage{bibentry}

\begin{document}

\twocolumn[{
\renewcommand\twocolumn[1][]{#1}
\maketitle
\vspace{-9mm}
\begin{center}
\centering
\captionsetup{type=figure}
\includegraphics[width=1\linewidth]{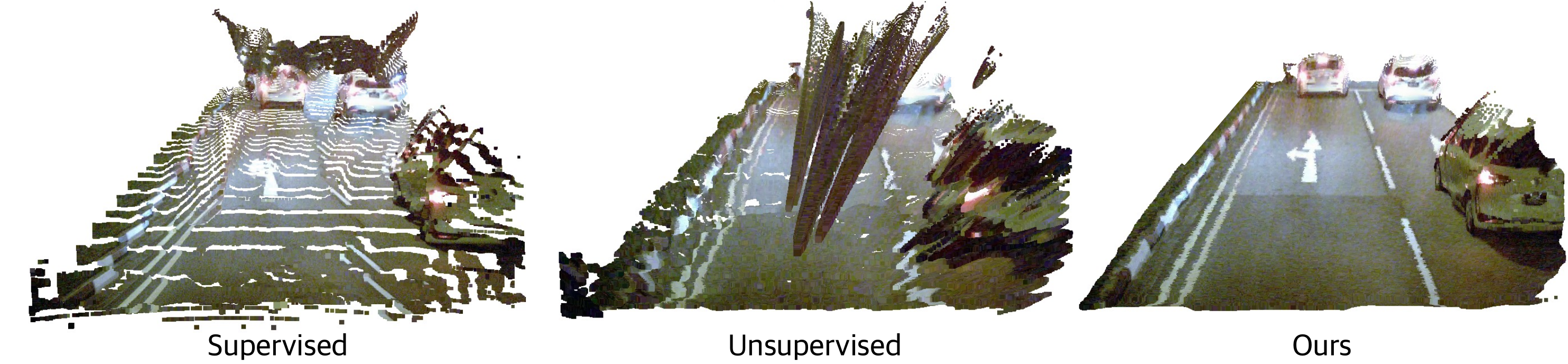}
   \caption{\textbf{3D-lifted depth completion results in out-of-domain cases.}
   Regardless of supervised~\cite{zhang2023completionformer} or unsupervised methods~\cite{wong2021unsupervised}, most depth completion models perform poorly on out-of-domain data.
   In contrast, our zero-shot depth completion method, which employs test-time alignment, consistently achieves robust results.
   In this example, the other models are trained on the KITTI Depth Completion dataset~\cite{uhrig2017sparsity}, while our zero-shot approach is not trained on any specific depth completion dataset. Both are tested on the nuScenes dataset~\cite{caesar2020nuscene}.
} 
\label{fig:teaser}
\end{center}
}]

\begin{abstract}
  Depth completion, predicting dense depth maps from sparse depth measurements, is an ill-posed problem requiring prior knowledge.
  Recent methods adopt learning-based approaches to implicitly capture priors, but the priors primarily fit in-domain data and do not generalize well to out-of-domain scenarios.
  To address this, we propose a zero-shot depth completion method composed of an affine-invariant depth diffusion model and test-time alignment.
  We use pre-trained depth diffusion models as depth prior knowledge, which implicitly understand how to fill in depth for scenes.
  Our approach aligns the affine-invariant depth prior with metric-scale sparse measurements, enforcing them as hard constraints via an optimization loop at test-time.
  Our zero-shot depth completion method
  demonstrates generalization across various domain datasets, achieving up to a 21\% average performance improvement over the previous state-of-the-art methods while enhancing spatial understanding by sharpening scene details.
  We demonstrate that aligning a monocular affine-invariant depth prior with sparse metric measurements is a proven strategy to achieve domain-generalizable depth completion without relying on extensive training data. Project page: \small{\emph{\url{hyoseok1223.github.io/zero-shot-depth-completion/}}}.
\end{abstract}

\vspace{-1mm}
\section{Introduction}
\label{sec:intro} 
Metric-scale dense depth provides precise spatial structure of a scene, crucial for physically accurate applications such as 3D scene understanding~\cite{jiyeon2024unidvps}, 3D reconstruction~\cite{choe2021volumefusion}, and robotic grasping~\cite{viereck2017learning}. 
This depth information is essential for achieving reliable and robust performance across real-world perception and interaction, where 
failures can lead to significant risks.
However, acquiring dense metric depth map in practical settings is challenging, as depth measurements captured by depth sensing approaches -- long-range sensors (\eg, LiDAR)~\cite{Ma2017SparseToDense} 
and SLAM/VIO systems~\cite{wong2021unsupervised} -- are sparse 
potentially leading to safety risks.
To complement this limitation, depth completion has been studied, which aims to complete the dense metric depth map from sparse measurements.

However, depth completion is an ill-posed problem requiring prior knowledge and additional cues, \eg, RGB images as guidance~\cite{Ma2017SparseToDense, hu2021penet, guidenet, qiu2019deeplidar}. 
Previous studies~\cite{park2020nonlocal, zhang2023completionformer, wong2021unsupervised, wang2023lrru} 
have focused on learning how to propagate sparse metric depth into a dense map according to the color or texture proximity.
They are trained with paired dense depth maps and corresponding RGB images to learn 
depth affinity as prior knowledge, where
the depth affinity represents the relationship between depth values in a scene
based on spatial and structural features.
Since previous methods~\cite{zhang2023completionformer, wong2021unsupervised} focused on
learning depth affinity within in-domain settings, 
they exhibit poor depth affinity
in out-of-domain scenarios (see \Fref{fig:teaser}).
To address this, \citet{park2024testtime} 
proposed a test-time adaptation method that fine-tunes part of a pre-trained depth completion model using sparse depth.
Nevertheless, This approach is less effective in out-of-domain scenarios due to the limited generalizability of the base depth completion model.

With the emergence of foundation models~\cite{caron2021emerging, rombach2022highresolution}, which learn comprehensive knowledge from large image data (referred to as image prior), these models have been frequently utilized as powerful prior to improve generalizability, enabling them to be applicable across diverse tasks and domains~\cite{lee2024dmp, yang2023diffusion, liu2023grounding}.
We bring this versatile capability to the depth completion problem.
In this regime, we propose 
zero-shot depth completion via a test-time alignment,
which is generalizable to any domain by leveraging the rich semantic and structural understanding of the foundation model.

Specifically, we use pre-trained monocular depth diffusion models \cite{ke2023repurposing, gui2024depthfm} as depth prior, 
demonstrating generalizability and facilitating high-quality depth estimation.
Most monocular depth estimation models~\cite{Ranftl2022midas, ke2023repurposing, depthanything, gui2024depthfm} operate in the affine-invariant depth space, where depth values are consistent up to offset and scale.
While this approach enables training on large-scale dataset with diverse scene contents and varying camera intrinsics~\cite{ke2023repurposing}, it inherently introduces scale ambiguity, making fully accurate monocular metric depth estimation to be considered infeasible~\cite{yin2023metric}.
Meanwhile, depth completion is free from scale ambiguity thank to sparse measurements of metric depths, but lacks generalizability and depth quality~\cite{park2024testtime}.
Motivated by these trade-offs, we align the affine-invariant depth prior with sparse measurements in the metric depth space, achieving generalizable and well-structured depth completion.
By performing this alignment at test time, we can complete the metric depth map from any pair of RGB and synchronized sparse depth data, \ie, zero-shot.
Figure~\ref{fig:teaser} illustrates the robustness of our method in the out-of-domain scenarios.

To this end, we propose a test-time alignment method that guides the reverse sampling process of the diffusion model
by incorporating optimization loops to enforce the given sparse depth as hard constraints. 
We also introduce a prior-based outlier filtering method to ensure reliable measurements and a new loss function to maintain the structural prior inherent in the depth prior.
Our method demonstrates superior generalization ability across various domain datasets~\cite{silberman2012nyu, mccormac2017scenenet, sun2020waymo, caesar2020nuscene}, including both indoor and outdoor environments.
Our contribution points are as follows:

\begin{itemize}
    \setlength\itemsep{0.3em}
    \item 
    We propose a novel zero-shot depth completion method that leverages foundation model prior to enhance domain generalization while capturing detailed scene structure.
    
    \item 
    We introduce a test-time alignment that uses sparse measurements as hard constraint to guide the diffusion sampling process, aligning with an affine-invariant depth prior.

    \item 
    We present a prior-based outlier filtering algorithm to improve the reliability of sparse measurements,
    enhancing the robustness of our method using 
    sparse depth guidance.    
\end{itemize}
\section{Related Work}
\para{Depth completion}
Depth completion is an ill-posed problem that aims to reconstruct unknown dense depth from observed
sparse depth measurements, with missing areas typically covering less than $5\%$ of an image for outdoor driving scenarios and $1\%$ for indoor scenarios~\cite{wong2020void}.
Since the success of deep learning, 
the problem 
has been addressed
by data-driven approaches that learn how to propagate sparse depth measurements
guided by the RGB images~\cite{wong2021unsupervised, park2020nonlocal}.
Prior studies~\cite{park2020nonlocal, lin2022dynamic, zhang2023completionformer} use affinity-based spatial propagation methods~\cite{liu2017learning, cheng2019learning}
to learn the relationship between dense depth and RGB pairs.
They learn how to propagate depth while preserving scene structure and boundaries.
This learning process requires large pairs of RGB images and dense depth maps, but acquiring these dense maps in real-world scenarios is costly due to dedicated sensor systems and requires careful data processing and curation.~\cite{uhrig2017sparsity, wong2020void}.
Depending on how to process data, domain discrepancies are introduced in each dataset, which makes depth completion models hard to generalize.

To mitigate these challenges arising from the lack of real data and domain gaps, unsupervised learning or domain adaptation methods have been proposed. 
Unsupervised methods~\cite{wong2021unsupervised, ma2018self, wong2020void} train a model 
with pairs of a RGB image and synchronized sparse depth without a dense depth map.
These methods exploit multi-view 
photometric consistency with multiple views to compensate for the lack of direct 3D supervision. 
As an alternative direction to mitigate lack of data and domain gaps, some works~\cite{wong2021scaffnet, lopezrodriguez2020project} is initially trained in a synthetic domain with supervised learning, followed by unsupervised training on real datasets as a way of
domain adaptation.
Different from these research, we tackle
the limitations by exploiting learned prior embeded in a foundation model. 
We use a pre-trained generative diffusion model that understands depth affinity, spatial detail, and scene context.
This strong prior from the foundation model further enables zero-shot generalization to any domain.

\para{Test-time Adaptation (TTA)}
Applying a model trained on a source domain to unseen test domains is crucial for generalization, especially in depth completion, where domain gaps arise from sensor variations, environmental conditions (\eg, weather changes), scene variety (\eg, driving locations), and depth ranges (\eg, indoor vs. outdoor). 
TTA methods~\cite{wang2021tent, wang2022continual, park2024testtime} address this by adapting models to unseen data.
However, they still suffer from domain gaps due to reliance on the source dataset, and often require additional training and continual adaptation, which may not be feasible in zero-shot scenarios.

With the emergence of foundation models, there has been a shift towards leveraging their prior knowledge for generalization across diverse tasks and domains~\cite{ jia2023dginstyle, liu2023grounding}.
As a generative foundation model, diffusion models are similarily employed as a generalizable priors.
To address the domain gaps in depth completion, we utilize a diffusion model that comprehends depth prior~\cite{ke2023repurposing, gui2024depthfm} by aligning it with sparse depth measurement 
using the proposed test time alignment method.
This approach effectively mitigates issues caused by domain gaps and enables depth completion in a zero-shot manner.


\begin{figure}[t]
\centering
\includegraphics[width=0.95\linewidth]
   {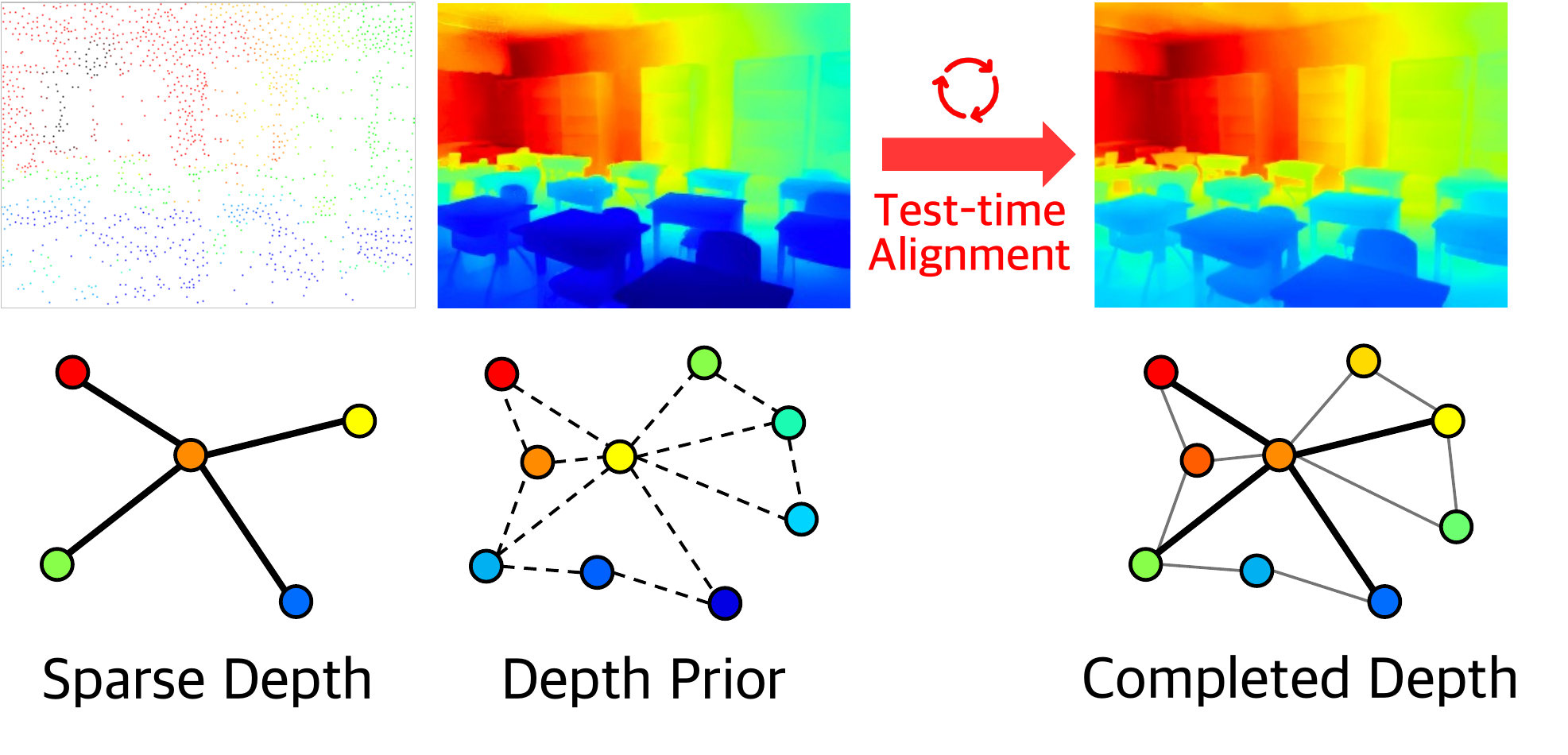}
   \setlength{\abovecaptionskip}{-2pt}
   \caption{\textbf{Illustration of our approach.}
    At test time, we align the depth affinity from the prior (dashed lines) with the sparse depth measurements as a hard constraint (bold lines). 
    This alignment propagates measurements across the scene to complete unobservable depth values.
   } 
\label{fig:concept}
\vspace{-3mm}
\end{figure}
\section{Method}
In this section, we introduce our zero-shot depth completion method, which leverages the depth prior~\cite{ke2023repurposing, gui2024depthfm}
derived from the foundation model~\cite{rombach2022highresolution}. This 
enables our method to be generalizable across any domain.
The core concept of our approach is to align the affine-invariant depth prior with sparse measurements on an absolute scale to complete the dense and well-structured depth map, as illustrated in \Fref{fig:concept}.

\subsection{Preliminary}
\label{sec:preliminary}
\para{Diffusion model and guided sampling}
Diffusion models~\cite{ho2020denoising, song2022denoising} aim to model data distribution $p(\mathbf{x})$ through iterative perturbation and restoration, known as forward and reverse processes.
This is represented by the score-based generative model~\cite{song2021scorebased}, learning the score function $\mathbf{s}_\theta$ parameterized by $\theta$ the gradient of the log probability density function with respect to the data, \ie, $\mathbf{s}_\theta(\mathbf{x}) =\nabla_{\mathbf{x}} \log p(\mathbf{x};\theta)$.
Score-based diffusion models 
estimate the score $\mathbf{s}_\theta(\mathbf{x}_t)$ at intermediate state $\mathbf{x}_t$ for timestep $t$ which defines a process.

For image generation and editing, diffusion models leverage the guidance function during the sampling process to adjust the output to the specific condition
~\cite{ho2022classifierfree, dhariwal2021diffusion}.
The guidance can be 
defined 
by any differentiable mapping output to guidance modality, as follows~\cite{bansal2024universal}: 
\begin{equation}
\label{eq:guide_sampling}
\hat{\mathbf{s}}_{\theta}(\mathbf{x}_t, t, \mathbf{y}) = \mathbf{s}_{\theta}(\mathbf{x}_t, t) + w \nabla_{\mathbf{x}_t} \mathcal{L}\left(f\left(\mathbf{x}_0\left(\mathbf{x}_t\right)\right),\mathbf{y}\right),
\end{equation}
where $w$ and $\mathbf{y}$ represent weight and guidance, respectively.
The function $f(\cdot)$ can be any differentiable function whose output can compute a loss $\mathcal{L}$ with guidance condition $\mathbf{y}$, and
$\mathbf{x}_0\left(\mathbf{x}_t\right)$ is obtained by using Tweedie's formula~\cite{efron2011tweedie}, 
which provides an approximation of the posterior mean. 
This guided sampling approach extends unconditional diffusion models to conditional ones without separate model training.

\para{Inverse problem}
The goal of an inverse problem is to determine an unknown variable from known measurement, often formulated as $\mathcal{A}(\mathbf{x})=\mathbf{y}$,
where $\mathcal{A}{:}\, \mathbb{R}^m {\rightarrow}  \mathbb{R}^n$ represents the known forward measurement operator, $\mathbf{y}\in \mathbb{R}^n$ and $\mathbf{x}\in \mathbb{R}^m$,
the measurement and the unknown variable, respectively.
When 
$m>n$, it becomes an ill-posed problem, requiring a prior to find 
solve a
Maximum A Posterior (MAP) estimation:
\begin{equation}
\label{eq:map}
    \argmax p(\mathbf{x}|\mathbf{y})\propto p(\mathbf{x}) p(\mathbf{y} | \mathbf{x}),
\end{equation}
where $p(\mathbf{x})$ represents our prior of the signal $\mathbf{x}$ and $p(\mathbf{y} | \mathbf{x})$ is likelihood measuring 
$\mathcal{A}(\mathbf{x})\approx\mathbf{y}$, \eg, $\|\mathbf{y} {-} \mathcal{A}(\mathbf{x})\|_2^2$.
By taking $-\log(\cdot)$ to \Eref{eq:map}, it can be  
formulated as an optimization problem
that regularizes the solution, ensuring that $\mathbf{x}$ follows the characteristics of the prior:
\begin{equation}
    \label{eq:inv_opt}
    \argmin_{\mathbf{x}} \left\|\mathbf{y} - \mathcal{A}\left(\mathbf{x}\right) \right\|_2^2 - \log p(\mathbf{x}).
\end{equation}
Also, given the gradient of $\log p(\mathbf{x}|\mathbf{y})$ in \Eref{eq:map} as
\begin{equation}
    \nabla_{\mathbf{x}}\log p(\mathbf{x}|\mathbf{y}) =  \nabla_{\mathbf{x}}\log p(\mathbf{x}) + \nabla_{\mathbf{x}}\log p(\mathbf{y} | \mathbf{x}),
\end{equation}
the prior term $\nabla_{\mathbf{x}}\log p(\mathbf{x})$ corresponds to the score $\mathbf{s}_{\theta}(\mathbf{x})$, which can be obtained by diffusion models.
Therefore, by simply adding the gradient of the likelihood term to the reverse sampling process, the inverse problem can be effectively solved while leveraging the diffusion prior~\cite{chung2023dps}
as follows:
\begin{equation}
    \label{eq:inv_sampling}
    \hat{\mathbf{s}}_{\theta}(\mathbf{x}_t, t, \mathbf{y}) = \mathbf{s}_{\theta}(\mathbf{x}_t, t) + w \nabla_{\mathbf{x}_t}\left\| \mathbf{y} - \mathcal{A}\left(\mathbf{x}_0\left(\mathbf{x}_t\right)\right) \right\|_2^2.
\end{equation}
This has an analogous form with \Eref{eq:guide_sampling}; thus, the inverse problem can be effectively tackled with the guided sampling.

With pre-trained image diffusion models, \eg, \citet{rombach2022highresolution}, as the score function $\mathbf{s}_{\theta}(\mathbf{x})$ and a prior, it provides 
powerful image prior across various tasks by its 
comprehensive semantic understanding and structural knowledge learned from a lot of images~\cite{wang2023exploiting, namekata2024emerdiff}.
\citet{ke2023repurposing} leverage this rich visual knowledge to achieve generalizable monocular depth estimation, resulting in high-quality outputs within an affine-invariant depth space. In our work, we exploit this depth diffusion model for computing the score as a depth prior.

\begin{figure*}[t]
\centering
   \includegraphics[width=0.9\linewidth]{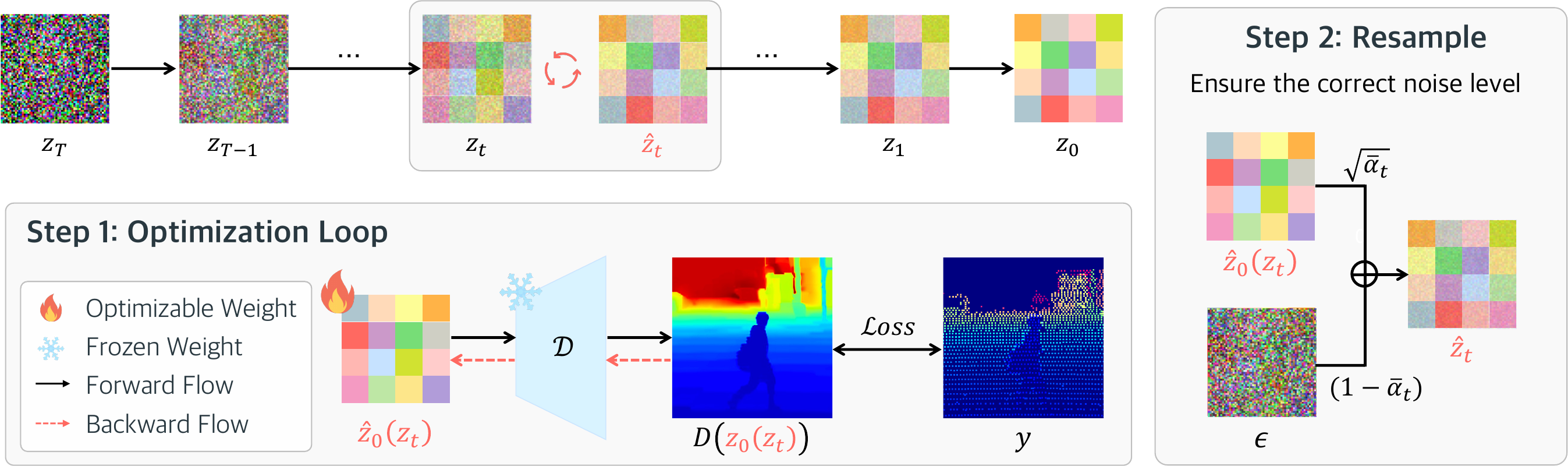}
   \caption{\textbf{Test-time alignment process.} 
   We incorporate a two-step hard alignment process into the reverse sampling process including an optimization loop and resample at regular intervals.
   We optimize \coral{$\mathbf{z}_0(\mathbf{z}_t)$} and remap it to \coral{$\hat{\mathbf{z}}_t$}.
   The latent is then decoded into depth, where the loss is measured against sparse depth.
   For visibility, the sparse depth points are enlarged.
   } 
\label{fig:opt_loop}
\vspace{-5mm}
\end{figure*}
\para{Problem formulation}
\label{sec:problem_form}
To leverage the prior knowledge, we formulate
a depth completion as an inverse problem that estimates unknown dense depth from 
observed sparse measurements.
$\mathbf{y}$ represents the observed sparse depth, 
$\mathbf{x}$ is the unknown dense depth, and 
$\mathcal{A}{:}\,\mathbb{R}^m{\rightarrow}\mathbb{R}^n$ is a binary measurement matrix of which entry  $[\mathcal{A}]_{ij}$ is $1$ if the entities $[\mathbf{y}]_i$ is measured from $[\mathbf{x}]_j$, $0$ otherwise. 
We follow \Eref{eq:inv_sampling}, where sparse depth serves as guidance.
We use the depth diffusion models \cite{ke2023repurposing, gui2024depthfm} extended from the latent diffusion model (LDM)~\cite{rombach2022highresolution} as prior, where
$\mathbf{x}$ is 
decomposed with the decoder $\mathcal{D}{:}\, \mathbf{z} \rightarrow \mathbf{x}$ as:
\begin{equation}
\label{eq:depth_guide_sampling}
\hat{\mathbf{s}}_{\theta}= \mathbf{s}_{\theta}(\mathbf{z}_t, t) + w {\nabla_{\mathbf{z}_t}}\left\| \mathbf{y} - \mathcal{A}\left(\mathcal{D}\left(\mathbf{z}_0\left(\mathbf{z}_t\right)\right)\right) \right\|_2^2,
\end{equation}
where $\mathbf{z}\in \mathbb{R}^{4\times H \times W}$ represents the latent of LDM but the decoder output $\mathbf{x}$ is treated as a flatten vector for convenience.

\subsection{Test-time Alignment with Hard Constraints}
\label{sec:opt_sampling}
Depth measurements obtained in practice are often sparse, unevenly distributed, and noisy. 
When the sparse measurements are used as guidance, the ill-posed nature of the problem, combined with the stochastic behavior of diffusion models, can lead to scores that produce undesirable solutions~\cite{kim2024regtext} and does not even guarantee that the estimation corresponds to the known sparse measurements. 
To deal with this, we propose a test-time alignment that incorporates the correction step 
to enforce the sparse measurement as harder constraints than encouraging guidance in a soft manner by \Eref{eq:depth_guide_sampling}.
This involves an optimization loop at regular intervals to enforce
measurement constraints as a correction step.
We further show the potential for uncertain solutions from the stochastic process in the supplementary material, illustrating why the alignment is necessary.

Additionally, we adopt $\mathbf{z}_0(\mathbf{z}_t)$ as optimizable variable.
Pre-trained diffusion models take input $\mathbf{z}_t$ aligend with the noise level at each timestep $t$.
However, directly optimizing $\mathbf{z}_t$  without considering input characteristics may lead to suboptimal results
\cite{chung2022improving, chung2023dps, chung2024dds}.
To address this, inspired by \citet{song2024solving}, we use $\mathbf{z}_0(\mathbf{z}_t)$ estimated from $\mathbf{z}_t$.
The optimization loop is formulated as:
\begin{equation}
    \label{eq:opt_loop}
    \hat{\mathbf{z}}_0(\mathbf{z}_t) = \argmin_{\mathbf{z}_0(\mathbf{z}_t)} \left\| \mathbf{y} - \mathcal{A}\left(\mathcal{D}\left(\mathbf{z}_0\left(\mathbf{z}_t\right)\right)\right) \right\|_2^2.
\end{equation}
Then, to ensure adherence to the correct noise level, the measurement-consistent $\hat{\mathbf{z}}_0(\mathbf{z}_t)$ is remapped to an intermediate latent $\hat{\mathbf{z}}_t$ by adding time-scheduled Gaussian noise, as expressed below:
\begin{equation}
    \label{eq:remap}
    p\left(\hat{\mathbf{z}}_{t} | \hat{\mathbf{z}}_0(\mathbf{z}_t)\right) = \mathcal{N}(\sqrt{\bar{\alpha}_{t}} ~ \hat{\mathbf{z}}_0(\mathbf{z}_t), (1 - \bar{\alpha}_{t}) I),
\end{equation}
\noindent where $\bar{\alpha}_{t} = \prod_{i=1}^t \alpha_i,$ and $\alpha_t$ is variance schedule at time $t$.

\begin{figure*}[t]
\centering
   \includegraphics[width=1.0\linewidth]{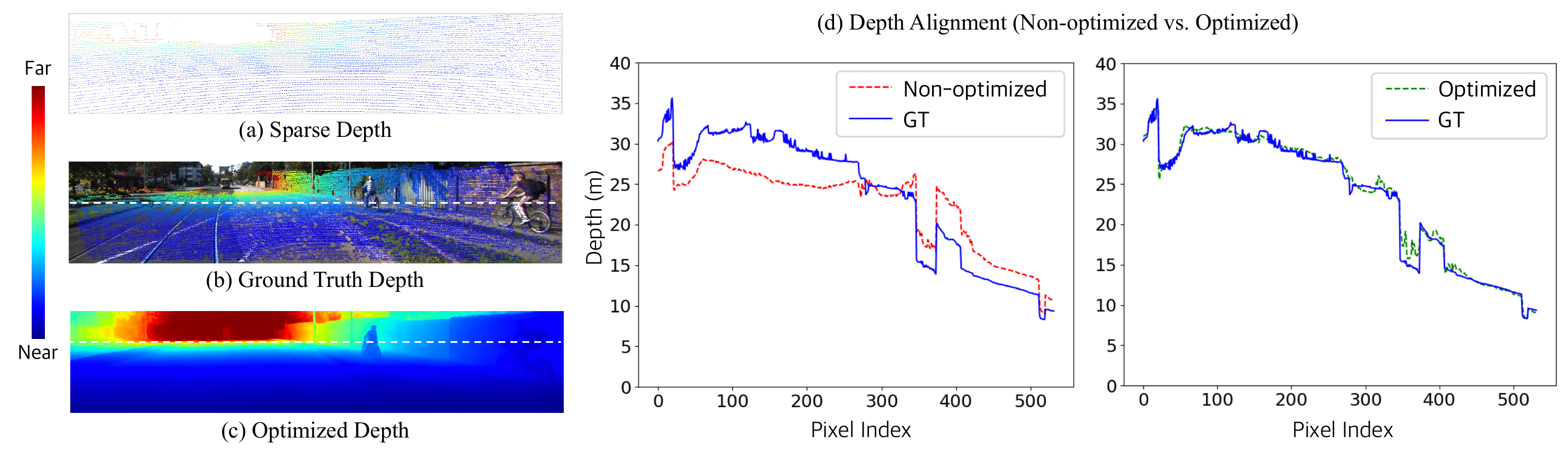}
   \caption{\textbf{Alignment with metric depth.} 
   We evaluate our method's effectiveness against ground truth (GT), accumulated semi-densely.
   We use only sparse depth (a) to align with actual metric depth values in complex scenes, ensuring a desirable solution.
   The white lines in (b), (c), and the x-axis of (d) represent pixel indices with valid depth points in a row of the GT.
   } 
\label{fig:depth_align}
\vspace{-3mm}
\end{figure*}

Since the score $\hat{\mathbf{s}}_{\theta}(\mathbf{z}_t, t)$ is directly added to the latent $\mathbf{z}_t$ at each step,
\Eref{eq:depth_guide_sampling} can be rewritten in terms of $\mathbf{z}_0(\mathbf{z}_t)$ with a modulated weight factor $\zeta$, as follows:
\begin{equation}
    \label{eq:depth_guide_sampling_latent}
    \hat{\mathbf{z}}_t = \mathbf{z}_t + \zeta \nabla_{\mathbf{z}_t} \left\| \mathbf{y} - \mathcal{A}\left(\mathcal{D}\left(\mathbf{z}_0(\mathbf{z}_t)\right)\right) \right\|_2^2.
\end{equation}
Here, \Eref{eq:depth_guide_sampling_latent} is replaced by the two-step process of \Eref{eq:opt_loop} and \Eref{eq:remap}, allowing our test-time alignment process to effectively achieve measurement-consistent desirable solutions.
Figure \ref{fig:opt_loop} illustrates the our test-time alignment process.
Figure~\ref{fig:depth_align} demonstrates how effectively our test-time alignment method estimates unseen depth areas by aligning sparse measurements with an affine-invariant depth prior. This result highlights the need for correction.
Examples of undesirable solutions and their corrected ones by our method are provided in the supplementary material.

Until now, in solving \Eref{eq:depth_guide_sampling}, we use an affine-invariant depth model for completing metric depths without special care. However, a natural question arises: ``\textit{Is the affine-invariant depth model compatible with estimating metric depths in our framework?}'' The following analysis shows that it may be sufficient.

\vspace{1mm}\noindent\textbf{Can we use an affine-invariant depth model for completing metric depths?}
Depth estimation models are often trained to estimate affine-invariant depth with scale and shift invariant loss to achieve generalizable performance~\cite{Ranftl2022midas, ke2023repurposing, eigen2014invariant}.
Thus, depth prior operates in the affine-invariant depth space, which does not directly correspond to the metric depth used in measurements.
Even though the given sparse metric depth is normalized between 0 and 1, 
their statistics including 
mean and variance
can differ, and the relationship between real metric depth and estimated affine-invariant depth is often 
non-linear (see 
the left of \Fref{fig:depth_align} (d)).
Therefore, to determine if 
\Eref{eq:depth_guide_sampling} can be used to solve this problem, we need to verify whether the normalized metric depth space lies within the data distribution generated by the diffusion model.

To confirm this, we conduct an empirical investigation through the following procedure: given $\tilde{\mathbf{x}}_0$, dense depth map estimated from the pre-trained depth completion model, we perform its reconstruction using an affine-invariant depth diffusion model.
This process involves sequentially encoding $\tilde{\mathbf{x}}_0$ to $\tilde{\mathbf{z}}_0$, 
then doing inversion by adding noise~\cite{song2022denoising}, which results in $\tilde{\mathbf{z}}_t$. 
Next, we perform reverse sampling, $\nabla_{\mathbf{z}_t}\log p(\tilde{\mathbf{z}}_t)$ with only the affine-invariant depth diffusion prior.
The reconstructed result achieves similar performance compared to the original one, $\tilde{\mathbf{x}}_0$, excluding encoding-decoding information loss. 
The details and results of the experiment are provided in the supplementary material.
This result suggests that the affine-invariant depth prior is sufficiently capable of handling the metric depth space, 
which corresponds to:
\begin{equation}
\label{eq:verify}
    \nabla_{\mathbf{z}_t}\log p(\tilde{\mathbf{z}}_t) \approx \nabla_{\tilde{\mathbf{z}}_t}\log p(\tilde{\mathbf{z}}_t).
\end{equation}
Thus, we just need to align this prior with metric depth cue validating using \Eref{eq:depth_guide_sampling} to solve ill-posed depth completion.

\subsection{Prior-based Outlier Filtering}
\label{sec:noise_filter}
Practical depth sensing methods often produce outliers, such as unsynchronized depth with RGB or see-through points~\cite{conti22confidence}), making sparse depth measurements unreliable.
This degrades the performance of methods relying on sparse depth supervision~\cite{wong2021unsupervised,wong2020void}.
We also use sparse depth measurement as supervision during test-time alignment, this makes the alignment process prone to divergence or slow convergence.
To address this,
we utilize data-driven depth prior~\cite{ke2023repurposing, gui2024depthfm}, which benefits from the more precise synchronization with RGB images and depth affinity.
To obtain outlier-free sparse points $\mathbf{y}^*$, we adopt a divide-and-conquer approach.
We define local segments based on depth affinity, grouping regions where relative depth values are similar within a spatially local area.
Within these segments, the depth distribution can be easily categorized into inliers and outliers, enabling us to effectively identify outliers.

Affine-invariant depth map $D_r$ is divided into local segments $S_i$, which are regions with a high probability of having similar depths with considering location. 
For this clustering 
we leverage the superpixel algorithm~\cite{achanta2012slic, li2015lsc}.
In each region, we perform linear least-square fitting to map affine-invariant depth to metric depth using sparse metric depth measurements $\mathbf{y}_i$.
However, since these sparse measurements are influenced by outliers, we use RANSAC~\cite{fischler1981ransac} to perform outlier-robust linear least-square fitting on points where noisy $\mathbf{y}$ intersects $S_i$ \ie, $\mathbf{y}_i \leftarrow S_i \cap \mathbf{y}$.
This allows us to estimate outlier-robust metric depth values $\hat{\mathbf{y}}_i$ in local regions $S_i$.
Then, points with significant deviations exceeding $\tau$ are identified as outliers and filtered out.
Our proposed filtering algorithm, based on monocular depth prior, is detailed in Algorithm~\ref{algorithm:1}.

\begin{algorithm}[t]
\caption{Prior-based outlier filtering algorithm.}
\label{algorithm:1}
\begin{algorithmic}[1]
\setlength{\itemsep}{0.15em}
    \State \textbf{Parameters:} Number of segments $N$, Filter threshold $\tau$
    \State \textbf{Input:} 
    Estimated relative depth $D_r$, Sparse metric depth $\mathbf{y}$, Set of sparse point locations $\Omega(\mathbf{y})$.
    \State \textbf{Output:} Set of reliable sparse point locations $\Omega(\mathbf{y}^*)$.
    \State $\{\Omega(S_i)\}_{i=1 \cdots N} \gets \text{SuperPixel}\left( D_r, N 
    \right)$ 
    \For{$i = 1$ \textbf{to} $N$}
        \State $\Omega(\mathbf{y}_i) \gets \Omega(\mathbf{y}) \cap \Omega(S_i)$
        \State $\hat{\mathbf{y}}_i \gets 
        {\text{RANSAC~Regressor}}(\mathds{1}_{\Omega(\mathbf{y}_i)} \odot D_r,~ \mathbf{y}_i)$ 
        \State $\Omega(\mathbf{y}_i^{*}) \gets |\hat{\mathbf{y}}_i-\mathbf{y}_i| > \tau$
    \EndFor
    \State \( \Omega(\mathbf{y}^*) \gets \bigcup_{i=1}^{N} \Omega(\mathbf{y}_i^{*}) \)
\end{algorithmic}
\end{algorithm}
\definecolor{tabfirst}{rgb}{1, 0.7, 0.7} 
\definecolor{tabsecond}{rgb}{1, 0.85, 0.7} 
\definecolor{tabthird}{rgb}{1, 1, 0.7} 
\newcommand{\mystrut}{\rule[-0.4ex]{0pt}{1.7ex}}
\newcommand{\highlight}[2]{\colorbox{#1}{\mystrut#2}}

\begingroup
\setlength{\tabcolsep}{12pt}
\begin{table*}[t]
\centering
\renewcommand{\arraystretch}{1.1} 
    \resizebox{0.9\linewidth}{!}{
    \begin{tabular}{m{2.5cm} cc cc cc cc }
    \toprule
    \multirow{3}[3]{*}{Method} & \multicolumn{4}{c}{Indoor} & \multicolumn{4}{c}{Outdoor} \\
    \cmidrule(lr){2-5} \cmidrule(lr){6-9} 
    & \multicolumn{2}{c}{NYUv2} & \multicolumn{2}{c}{SceneNet}
    & \multicolumn{2}{c}{Waymo} & \multicolumn{2}{c}{nuScenes} \\
     \cmidrule(lr){2-3} \cmidrule(lr){4-5} \cmidrule(lr){6-7} \cmidrule(lr){8-9}
    & RMSE & MAE & RMSE & MAE & RMSE & MAE & RMSE & MAE \\
    \midrule 
     Pre-trained 
     & 0.446 & 0.189 & 0.443 & 0.173
     & 2.821 & 1.514
     & 3.998 & 1.967 \\ 
     BNAdapt
     & 0.410 & 0.189 & 0.446 & 0.176
    & 2.194 & \cellcolor{tabthird}1.122
    & 1.801 & 0.828\\ 
     CoTTA
     & 0.376 & 0.147 & 0.405 & 0.136
     & 2.652 & 1.227
     & 2.668 & 1.222 \\ 
     ProxyTTA
     & \cellcolor{tabthird}0.203 & \cellcolor{tabthird}0.095 & \cellcolor{tabthird}0.357 & \cellcolor{tabthird}0.125
    & \cellcolor{tabthird}2.178 & \cellcolor{tabfirst}0.971
    & \cellcolor{tabthird}1.755 & \cellcolor{tabthird}0.799\\ 
    \midrule
    Ours~(+Marigold)
    & \cellcolor{tabsecond}0.149 & \cellcolor{tabfirst}0.059 & \cellcolor{tabsecond}0.207 & \cellcolor{tabsecond}0.099 
    & \cellcolor{tabfirst}2.115 & \cellcolor{tabsecond}1.121
    & \cellcolor{tabfirst}1.561 & \cellcolor{tabfirst}0.561\\
    Ours~(+DepthFM)
    & \cellcolor{tabfirst}0.145 & \cellcolor{tabsecond}0.077 & \cellcolor{tabfirst}0.178 & \cellcolor{tabfirst}0.081 
    & \cellcolor{tabsecond}2.162 & 1.133
    & \cellcolor{tabsecond}1.622 & \cellcolor{tabsecond}0.618\\
    \bottomrule
    \end{tabular}
    }
\caption{\textbf{Quantitative comparison of generalizable performance.}
We evaluate the generalizability of our method by comparing it with test-time adaptation methods across various domain datasets.
In this table, the pre-trained depth completion model is CostDCNet~\cite{kam2022costdcnet}, trained on KITTI DC for outdoor and VOID for indoor adaptation.
It is used for each adaptation method---BNAdapt~\cite{wang2021tent}, CoTTA~\cite{wang2022continual}, and ProxyTTA~\cite{park2024testtime}---excluding ours, for adapting to each domain.
The first best is marked in \highlight{tabfirst}{red}, the second in \highlight{tabsecond}{orange}, and the third in \highlight{tabthird}{yellow}.
}
\label{tab:generalization}
\vspace{-5pt}
\end{table*}
\endgroup

\subsection{Losses}
\label{sec:losses}
Our objective for optimization includes sparse depth consistency loss and regularization terms: a local smoothness loss to preserve depth prior and a new relative structure similarity loss to maintain structural prior inherent in depth prior.

\para{Sparse depth consistency}
Given the sparse depth measurement $y$, it ensures consistency with the metric depth.
To effectively integrate the observed measurements with affine-invariant depth prior and mitigate potential uncertainties, we employ $L_{1}$ loss as follows:
\begin{equation}
    \resizebox{0.59\hsize}{!}{$
    \mathcal{L}_{depth} = \scalebox{1.4}{$\frac{1}{|\Omega(\mathbf{y})|}$} \sum\limits_{\Omega(\mathbf{y})} |\mathbf{y}- \mathcal{A}(\hat{D})|,
    $}
\end{equation}
where $\mathcal{A}$ is the operation that Hadamard product with the zero-one mask $\mathds{1}_{\Omega(\mathbf{y})}$ and $\hat{D}$ represents completed depth.

\para{Local smoothness}
Using only sparse depth guidance risks losing the prior knowledge inherent in pre-trained depth diffusion models~\cite{ke2023repurposing, gui2024depthfm}, such as the property of depth which is locally smooth.
To mitigate this, we introduce a regularization term that enforces smoothness by applying the $L_1$ norm to gradients in both the $X$ and $Y$ directions, with reduced gradient weights near edges to prevent over-smoothing.
The loss function is defined as follows:
\begin{equation}
    \resizebox{0.89\hsize}{!}{$
    \mathcal{L}_{smooth} =\scalebox{1.5}{$\frac{1}{|\Omega|}$} \sum\limits_{c\in\Omega} \lambda_X(c) |\partial_X \hat{D}(c)| + \lambda_Y(c) |\partial_Y \hat{D}(c)|,
    $}
\end{equation}
where $\lambda_X(c)=e^{-|\partial_X I(c)|}$, $\lambda_Y(c)=e^{-|\partial_Y I(c)|}$, and $c \in \Omega$ represents the set of all pixel locations~\cite{park2024testtime}.
However, using only these loss functions may dilute the structural prior in the pre-trained depth diffusion model, which is key for detail sharpness. 

\para{Relative Structure Similarity}
To address this, we design a new structure regularization term that transfers structure from the depth estimated by an off-the-shelf model to regularize overly smooth structures.
Inspired by the structure similarity (SSIM) loss~\cite{wang2004ssim}, we propose the Relative Stucture Similarity (R-SSIM) loss, designed to transfer structure across domains.
This loss is derived from SSIM by dropping the luminance term, which relies on absolute values:
\begin{equation}
    \mathcal{L}_{r-ssim}(d_1,d_2) = 1-\frac{2\sigma_{d_1d_2} + C}{\sigma_{d_1}^2 + \sigma_{d_2}^2 + C},
\end{equation}
where $d_1$ and $d_2$ represent spatial information in different domains, $C$ is a constant, and $\sigma$ denotes the normalized standard deviation of pixel values.
Here, $d_1$ is the relative depth map, and $d_2$ is the estimated complete depth map (or vice versa).
The key point is that these domains may differ in pixel value ranges and statistics. 

\vspace{1.5mm}
\noindent Our comprehensive loss function is as follows:
\begin{equation}
    \mathcal{L} = \mathcal{L}_{depth} + \lambda_{smooth} \mathcal{L}_{smooth} + \lambda_{r-ssim} \mathcal{L}_{r-ssim},
\end{equation}
where $\lambda_{smooth}$ and $\lambda_{r-ssim}$ are regularization weights.


\begin{table}[t]
\centering
\renewcommand{\arraystretch}{1.3} 
    \resizebox{1.0\linewidth}{!}{
    \begin{tabular}{m{3.6cm} ccc}
    \toprule
     Base Model &  \makecell{Inference time} &RMSE & MAE\\
    \midrule
    Marigold ~(\textbf{50} steps) & 101s & 1.413 & 0.397\\
    DepthFM (\textbf{2} steps) & 31s & 1.499 & 0.377\\
    DepthFM (\textbf{1} step)  & 16s & 1.601 & 0.428\\
    \bottomrule
    \end{tabular}
    }
   \vspace{-3pt}
\caption{\textbf{Efficiency evaluation on the KITTI validation set.} Inference time of our method is measured as base models~\cite{ke2023repurposing, gui2024depthfm} with varying sampling 
}
\label{tab:efficiency}
\vspace{-5pt}
\end{table}

\section{Experiments}
In this section, we demonstrate the effectiveness of our prior-based depth completion method in indoor (NYUv2~\cite{silberman2012nyu}, SceneNet~\cite{mccormac2017scenenet}, VOID~\cite{wong2020void}) and outdoor (Waymo~\cite{sun2020waymo}, nuScenes~\cite{caesar2020nuscene}, KITTI DC~\cite{uhrig2017sparsity}) scenarios, through both quantitative and qualitative evaluations. 
For evaluation, we use the Root Mean Squared Error (RMSE) and Mean Absolute Error (MAE), both standard metrics in depth completion where lower values indicate better performance.
The results are reported in meters.
Further details are provided in the supplementary material.

\subsection{Domain Generalization}
\label{sec:exp_domain}
\begin{figure*}[t]
\centering
    \includegraphics[width=0.95\linewidth]{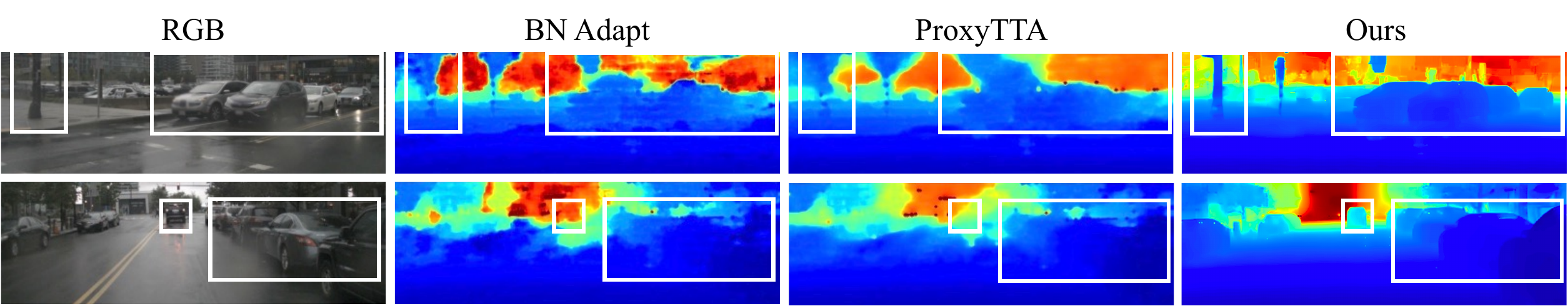}
   \vspace{-3pt}
   \caption{\textbf{Qualitative comparison on the nuScenes test set.}
    In outdoor scenarios, our test-time alignment method performs robustly even under extreme weather conditions, clearly identifying critical elements such as vehicles and signs.
       } 
\label{fig:qualitative_adapt_outdoor}
\vspace{-1mm}
\end{figure*}

\Tref{tab:generalization} summarizes the domain generalization performance of our method and previous test-time adaptation methods~\cite{wang2021tent,wang2022continual, park2024testtime} on indoor (NYU, SceneNet) and outdoor (Waymo, nuScenes).
Across various datasets, our prior-based approach consistently achieves the best or second-best performance.
 Notably, unlike test-time adaptation methods relying on pre-trained depth completion models in metric depth space, our method operates in affine-invariant depth space while achieving impressive performance.
Additionally, we demonstrate the model generality of our method by applying it to two depth diffusion models, Marigold~\cite{ke2023repurposing} and DepthFM~\cite{gui2024depthfm}, as shown in \Tref{tab:generalization}.
\Tref{tab:efficiency} further presents the inference time of our method across base models and sampling steps, demonstrating its potential for improving efficiency with minimal performance.
We also observe that our method captures details on the scene, reflecting true performance and demonstrating robust domain generalization as shown in~\Fref{fig:qualitative_adapt_outdoor} and \ref{fig:qualitative_adapt_indoor}.
We provide additional qualitative results in supplementary material.

\begin{figure*}[t]
\centering
    \includegraphics[width=0.95\linewidth]{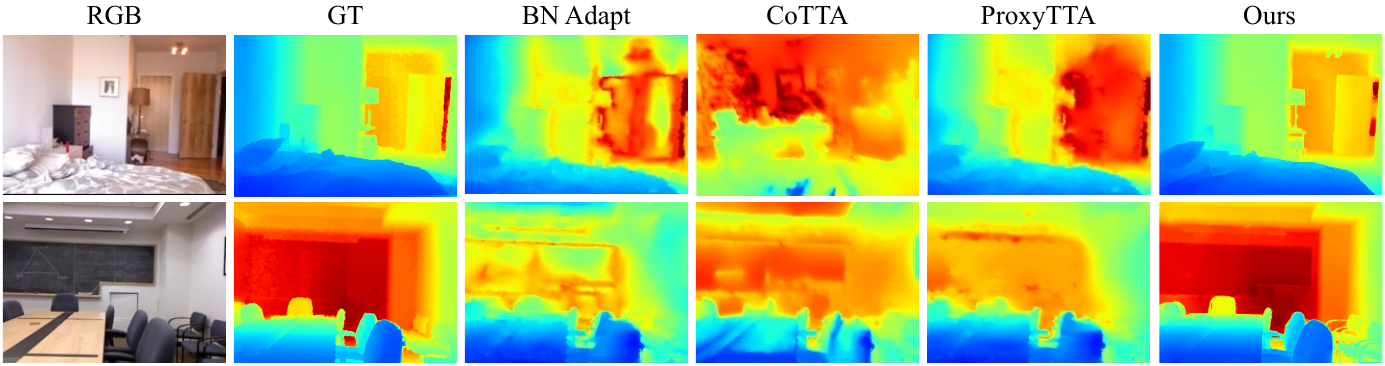}
   \vspace{-3pt}
   \caption{\textbf{Qualitative comparison on the NYU test set.}
   In indoor scenarios, our test-time alignment method accurately captures scene structures (\eg, chairs) compared to the existing test-time adaptation methods.
   } 
\label{fig:qualitative_adapt_indoor}
\vspace{-2mm}
\end{figure*}

In the outdoor datasets, the ground truth is obtained by accumulating LiDAR points after removing those corresponding to moving objects, which can lead to variations in the ground truth. 
For a more reliable benchmark, we use the ground truth provided by \citet{park2024testtime} for the Waymo and by \citet{huang2022pcacc} for the nuScenes.
In the supplementary material, we discuss in detail the differences in ground truth acquisition methods and their impact on the performance of depth completion methods.

\begin{table}[t]
\centering

\renewcommand{\arraystretch}{0.95} 
    \resizebox{1.0\linewidth}{!}{
    \begin{tabular}{m{4cm} c c cc}
    \toprule
    \multirow{2}[2]{*}{Method} & 
    \multirow{2}[2]{*}{\makecell{$N$-shot Scenario}} & \multirow{2}[2]{*}{RMSE} & \multirow{2}[2]{*}{MAE} \\
    & & & \\ [2pt]
    \midrule
    \addlinespace[4pt] 
    VPP4DC & 0  & 0.247 & 0.077\\ [4pt]
    DepthPrompting & 1  & 0.358 & 0.206\\ [2pt]
     & 10  & 0.220 & 0.101\\ [4pt]
    UniDC & 1  & 0.210 & 0.107 \\ [2pt]
     & 10  & 0.166 & 0.079 \\ [4pt]
     \midrule 
     \addlinespace[4pt] 
    Ours (+Marigold) & 0  & 0.149 & \textbf{0.059} \\[2pt]
    Ours (+DepthFM) & 0  & \textbf{0.145} & 0.077\\ [2pt]
    \bottomrule
    \end{tabular}
    }
   \vspace{-3pt}
\caption{\textbf{Quantitative comparison with depth-prior-based methods  on the NYU test set.} We compare our method with zero- and few-shot approaches leveraging various depth foundation models.
}
\label{tab:depthprior}
\vspace{-15pt}
\end{table}

\subsection{Comparison with Depth-Prior-Based Methods}
We compare our depth-prior-based method, which leverages depth diffusion models~\cite{ke2023repurposing, gui2024depthfm}, with other depth completion methods utilizing depth foundation models.
Each method relies on different depth foundation models: VPP4DC~\cite{bartolomei2024vpp4dc} employs a stereo matching network~\cite{lipson2021raft}, DepthPrompting~\cite{park2024depthprompting} utilizes ResNet34~\cite{he2016resnet} to extract depth features~\cite{lu2020depth, qiu2019deeplidar}, and UniDC~\cite{park2024unidc} leverages DepthAnything~\cite{depthanything}. 
\Tref{tab:depthprior} shows the effectiveness of our method leveraging depth diffusion models.

\subsection{Comparison with Unsupervised Methods}
\begin{table*}[t]
\centering
\renewcommand{\arraystretch}{1.1} 
    \resizebox{0.9\linewidth}{!}{
    \begin{tabular}{m{3.6cm} ccc cc cc }
    \toprule
    \multirow{2}[2]{*}{Method} & \multicolumn{3}{c}{Features} & \multicolumn{2}{c}{KITTI DC} & \multicolumn{2}{c}{VOID} \\
    \cmidrule(lr){2-4} \cmidrule(lr){5-6} \cmidrule(lr){7-8} 
    & \makecell{Sparse Depth \\ Supervision} & 
    \makecell{Photometric \\ Consistency Loss} &
    \makecell{In-domain \\ Training} 
    & RMSE & MAE & RMSE & MAE \\
    \midrule 
    Self-S2D & {\textcolor{green(ncs)}{\cmark}}  &
    {\textcolor{green(ncs)}{\cmark}} (two-view)
    & 
    {\textcolor{green(ncs)}{\cmark}} 
    & 1.384 & 0.358 & 0.243 & 0.178\\ 
    VOICED & {\textcolor{green(ncs)}{\cmark}}
    &
    \phantom{--}{\textcolor{green(ncs)}{\cmark}} (multi-view)
    & 
    {\textcolor{green(ncs)}{\cmark}}  
    &
    1.230 & 0.308 & 0.169 & 0.085\\ 
    ScaffNet & {\textcolor{green(ncs)}{\cmark}}

    &
    \phantom{--}{\textcolor{green(ncs)}{\cmark}} (multi-view)
    & 
    {\textcolor{green(ncs)}{\cmark}} 
    &
    1.182 & 0.286 & 0.119 & 0.059\\ 
    KBNet & {\textcolor{green(ncs)}{\cmark}} 

    &
    \phantom{--}{\textcolor{green(ncs)}{\cmark}} (multi-view)
    & 
    {\textcolor{green(ncs)}{\cmark}} 
    &
    1.126 & 0.260 & 0.095 & 0.039\\ 
    SPTR & {\textcolor{green(ncs)}{\cmark}} 
    &
    \phantom{--}{\textcolor{green(ncs)}{\cmark}} (multi-view)
    & 
    {\textcolor{green(ncs)}{\cmark}} 
    &
    1.111 & 0.254 & 0.091 & 0.040\\ 
    \midrule
    Ours w/ ~Our Filtering & & & & 1.413 & 0.397 & 0.111& 0.044\\ 
    Ours w/ ~Manual Filtering & \multirow{-2}{*}{\textcolor{green(ncs)}{\cmark}} & \multirow{-2}{*}{\phantom{-}\textcolor{red}{\xmark} (monocular)} & \multirow{-2}{*}{\textcolor{red}{\xmark}} & 1.198 & 0.287 & 0.112& 0.045\\ 
    \bottomrule

    \end{tabular}
    }
   \vspace{-3pt}
\caption{\textbf{Quantitative comparison with unsupervised methods.}
Despite weaker settings, our method performs comparably to unsupervised methods (Self-S2D~\cite{ma2018self}, VOICED~\cite{wong2020void}, ScaffNet~\cite{wong2021scaffnet}, KBNet~\cite{wong2021unsupervised}, and SPTR~\cite{sptr}) when sparse depth, \ie the supervision signal, is reliable.
 To demonstrate this, we ablate two filtering methods: our prior-based filtering and manual filtering, which is the outlier filtering method suggested by each benchmark.
 In this table, our method uses Marigold~\cite{ke2023repurposing} as the base model.}
\label{tab:unsup_comparison}
\vspace{-2mm}
\end{table*}

\begin{figure}[t]
\centering
   \includegraphics[width=0.95\linewidth]{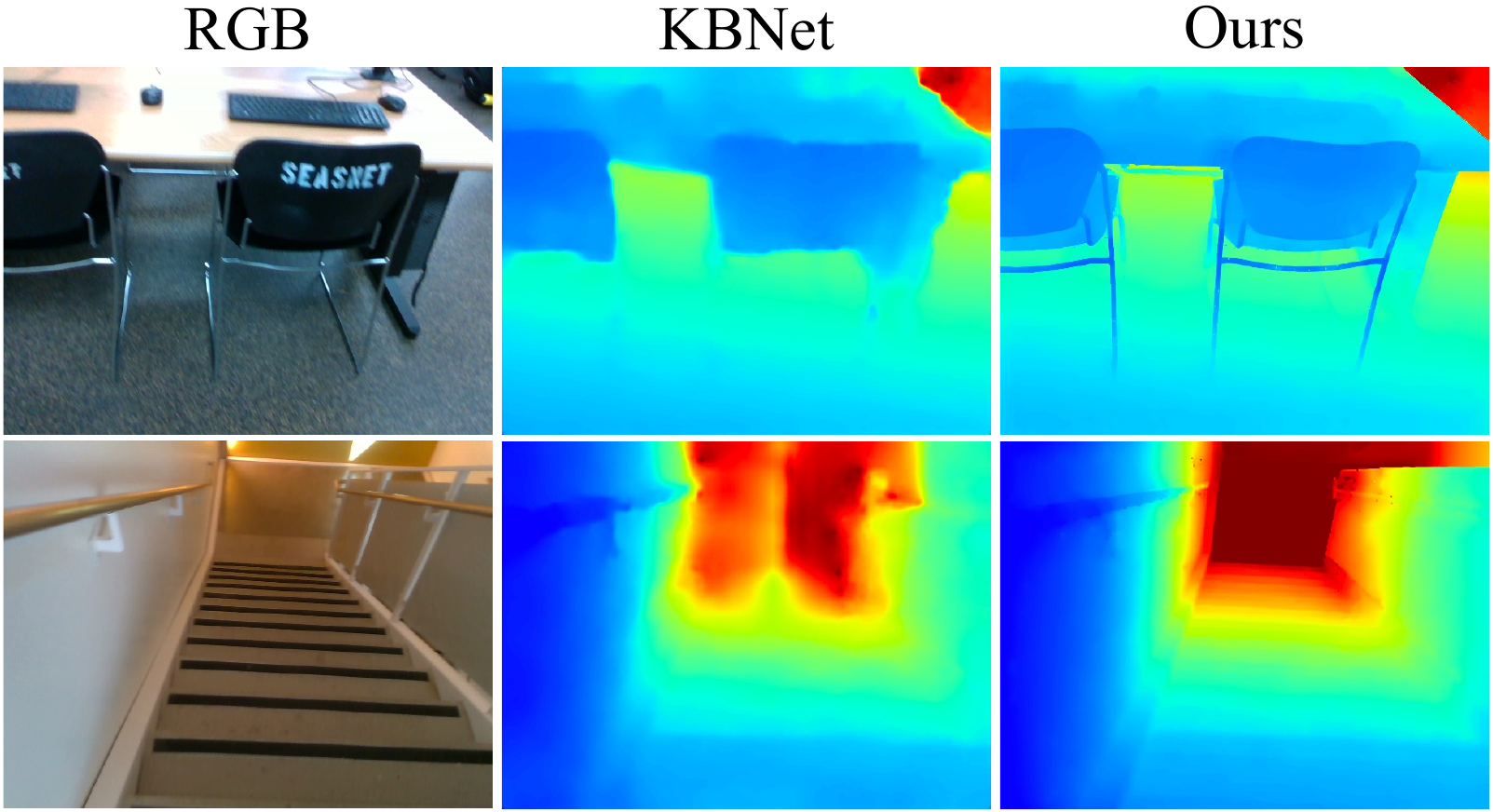}
   \vspace{-3pt}
   \caption{\textbf{Qualitative comparison on theVOID test set.}
    Compared to the state-of-the-art unsupervised method KBNet~\cite{wong2021unsupervised}, which uses multi-view photometric consistency, our prior-based approach better preserves scene structures and details using only monocular input.
   } 
\label{fig:unsup_in}
\vspace{-4mm}
\end{figure}

We compare our zero-shot depth completion method with unsupervised methods~\cite{wong2021unsupervised, ma2018self, wong2021scaffnet} trained on the split training dataset of each benchmark, \ie, in-domain training.
As shown in~\Tref{tab:unsup_comparison}, our method demonstrates favorable performance without dense depth data, multi-view, and in-domain training on KITTI DC and VOID. 
Additionally, our method achieves comparable performance when adopting manual filtering, that is, the outlier filtering method suggested by each benchmark.
Figure~\ref{fig:unsup_in} shows qualitative results of ours and unsupervised methods. 
Our method achieves higher-fidelity depth completion, preserving the depth affinity better than other unsupervised methods.

\subsection{Ablation Studies}

\begin{figure}[!ht]
\centering
\begin{minipage}{\linewidth}
    \centering
    \renewcommand{\arraystretch}{1.3} 
    \resizebox{1.0\linewidth}{!}{
        \begin{tabular}{ccc  cc cc}
        \toprule
        \multirow{2}[2]{*}{\makecell{Sampling \\ Method}} &\multirow{2}[2]{*}{\makecell{R-SSIM \\ Loss}} & \multirow{2}[2]{*}{\makecell{Outlier \\ Filtering}} & \multicolumn{2}{c}{KITTI DC} & \multicolumn{2}{c}{VOID} \\
        \cmidrule(lr){4-5} \cmidrule(lr){6-7} 
         & & & RMSE & MAE & RMSE & MAE\\
        \midrule
        Naïve & & & 3.514 & 1.942 & 0.199 & 0.130\\
        Guided & & & 2.113 & 0.801 & 0.210 & 0.138\\ 
        Ours & & & 1.610 & 0.406 & 0.125 & 0.046 \\
        Ours & {\cmark} & & 1.502 & 0.409 & 0.111 & 0.044 \\
        Ours & {\cmark} & {\cmark} & 1.413 & 0.397 & 0.112 & 0.045\\
        \bottomrule
        \end{tabular}
    }
    \vspace{-3pt}
    \caption{\textbf{Ablation studies.} We ablate our proposed methods including test-time alignment, R-SSIM loss, and prior-based outlier filtering, to demonstrate their effectiveness.}
    \label{tab:ablation_method}
\end{minipage}

\vspace{10pt} 

\begin{minipage}{\linewidth}
    \centering
    \includegraphics[width=0.95\linewidth]{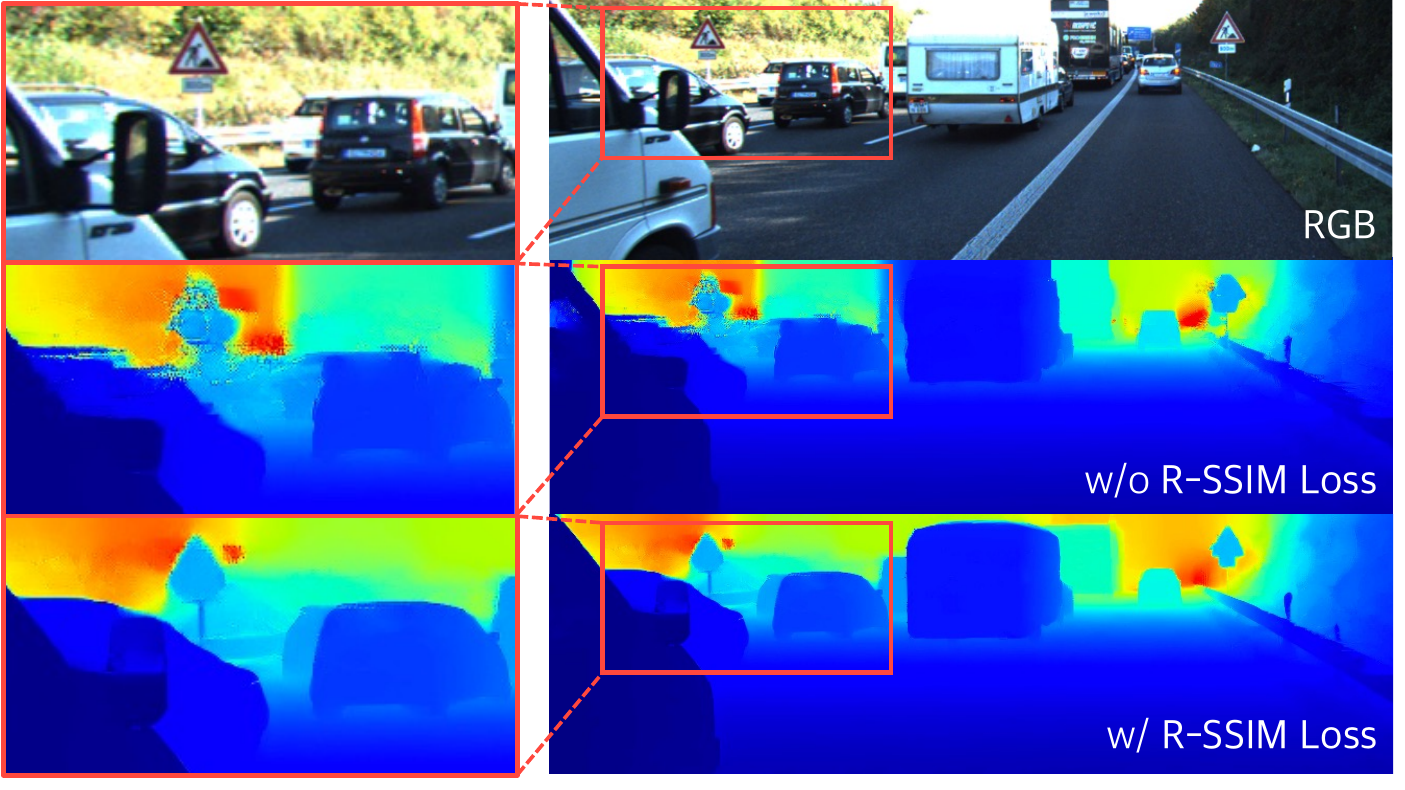}
    \vspace{-3pt}
    \caption{\textbf{Qualitative ablation of R-SSIM loss.} This structural regularization sharpens details in areas such as signposts and car shapes.} 
    \vspace{-10pt}
    \label{fig:r-ssim}
\end{minipage}
\end{figure}
\vspace{-5pt}
\Tref{tab:ablation_method} shows ablation studies to assess the efficacy of the test-time alignment method, R-SSIM loss, and outlier filtering algorithm.
The ablation studies are conducted on both indoor (VOID) and outdoor (KITTI DC) datasets.
Compared to other sampling methods, \ie, no guidance and the guided sampling~\cite{bansal2024universal}, the proposed test-time alignment method brings significant performance gain. The R-SSIM loss further enhances the performance and has a remarkable effect on preserving depth affinity.
The prior-based outlier filtering is more effective on the outdoor dataset 
than on the indoor dataset, as the sparse depth in the indoor dataset consists of
reliable points sampled from the ground truth.
We also qualitatively ablate the performance of the R-SSIM loss as shown in \Fref{fig:r-ssim}, highlighting how it effectively regularizes diffusion structural prior, leading to sharpen details.

\section{Conclusion}

We propose a novel prior-based zero-shot depth completion method, the first study demonstrating the importance of monocular depth prior knowledge in addressing the challenge of domain shifts.
Our test-time alignment approach ensures that the completed depth map remains consistent with sparse measurements while incorporating structural depth affinity of the scene derived from the depth prior.
This prior-based approach enhances the performance of depth completion across various domains, capturing the context of the scene.
We believe this work marks a significant step toward generalizable depth completion and our exploration of leveraging prior knowledge will inspire future work.

\para{Limitation}
Our zero-shot depth completion is the first work to use monocular depth foundation model priors for generalizable depth completion, but it adopts the standard guided sampling approach in latent diffusion models, which may be slow to process.
As a next step, accelerating this process building upon the recent advancements in the acceleration of diffusion model naïve~\cite{song2023consistency} and guided sampling~\cite{chung2022come} could be promising directions.


\vspace{-1mm}
\section*{Acknowledgement}
We thank the members of AMILab~\cite{ami} for their helpful discussions and proofreading. This work was supported by the RideFlux and Institute of Information \& communications Technology Planning \& Evaluation (IITP) grant funded by the Korea government(MSIT) (Development of Artificial Intelligence Technology for Self-Improving Competency Aware Learning Capabilities; No.RS-2022-II220124, Artificial Intelligence Innovation
Hub; No. 2019-0-01906, Artificial Intelligence Graduate
School Program(POSTECH)).

\newpage 
\bibliography{aaai25}

\appendix
\newpage

\setcounter{figure}{0}
\setcounter{table}{0}

\renewcommand{\thefigure}{S\arabic{figure}}
\renewcommand{\thetable}{S\arabic{table}}

\maketitlesupplementary

In this supplementary material, we present the details of experimental settings and additional experiments.  

\noindent\rule{\linewidth}{0.2pt}
\section*{Contents}
\vspace{2mm}
\subsection*{A. Experiment Setting and Details}
    \hspace*{1.5em}{{\ \ \ \textbf{A.1} \ \ \ Test-Time Alignment Details in Our Method}} \\
    \hspace*{1.5em}{{\ \ \ \textbf{A.2} \ \ \ Dataset Configurations}} \\
    \hspace*{1.5em}{{\ \ \ \textbf{A.3} \ \ \ Loss Functions}} \\

\subsection*{B. Additional Experiments}
    \hspace*{1.5em}{\ \ \ \textbf{B.1} \ \ \ Sensitivity of the Ground Truth Processing} \\
    \hspace*{1.5em}{\ \ \ \textbf{B.2} \ \ \ of nuScenes Benchmark} \\
    \hspace*{1.5em}{\ \ \ \textbf{B.3} \ \ \ Analysis of Test-Time Alignment Method} \\
    \hspace*{1.5em}{\ \ \ \textbf{B.4} \ \ \ Analysis of Outlier Filtering Method} \\

\subsection*{C. Additional Qualitative Results}
\noindent\rule{\linewidth}{0.2pt}

\section{Experiment Setting and Details}
\label{sec:dataset}
In this section, we provide the details of zero-shot depth completion via test-time alignment and dataset configuration of the divese test datasets.

\subsection{Test-Time Alignment Details in Our Method}
\noindent When we use Marigold~\cite{ke2023repurposing} for the affine-invariant depth diffusion model, our detailed settings are described below.
In the test-time alignment process, optimization starts after the first third of the total 50 reverse sampling steps and is performed every 5 steps thereafter.
Each optimization loop runs for 200 iterations. 
When we use DepthFM~\cite{gui2024depthfm} for the affine-invariant depth diffusion model, our detailed settings are described below.
DepthFM generally takes 1-2 steps for generative sampling acceleration. In the test-time alignment process, our optimization loop operates at all sampling steps. We present the results for each number of sampling steps in the main paper.

\noindent We set the weights of loss function, $\lambda_{smooth}$ and $\lambda_{r-ssim}$, to 0.2 and 0.3, respectively, and adjust them according to the dataset.
For high-resolution image data, such as from Waymo (1920x1280)~\cite{sun2020waymo} and nuScenes (1600x900)~\cite{caesar2020nuscene}, we optimize using $2\times$ downsampled images and then upsample them via bilinear interpolation. 
For our prior-based outlier filtering method, we segment superpixels into 200 segments.

\begin{figure*}[p]
\centering
\begin{minipage}{\textwidth}
   \centering
   \includegraphics[width=0.9\linewidth]{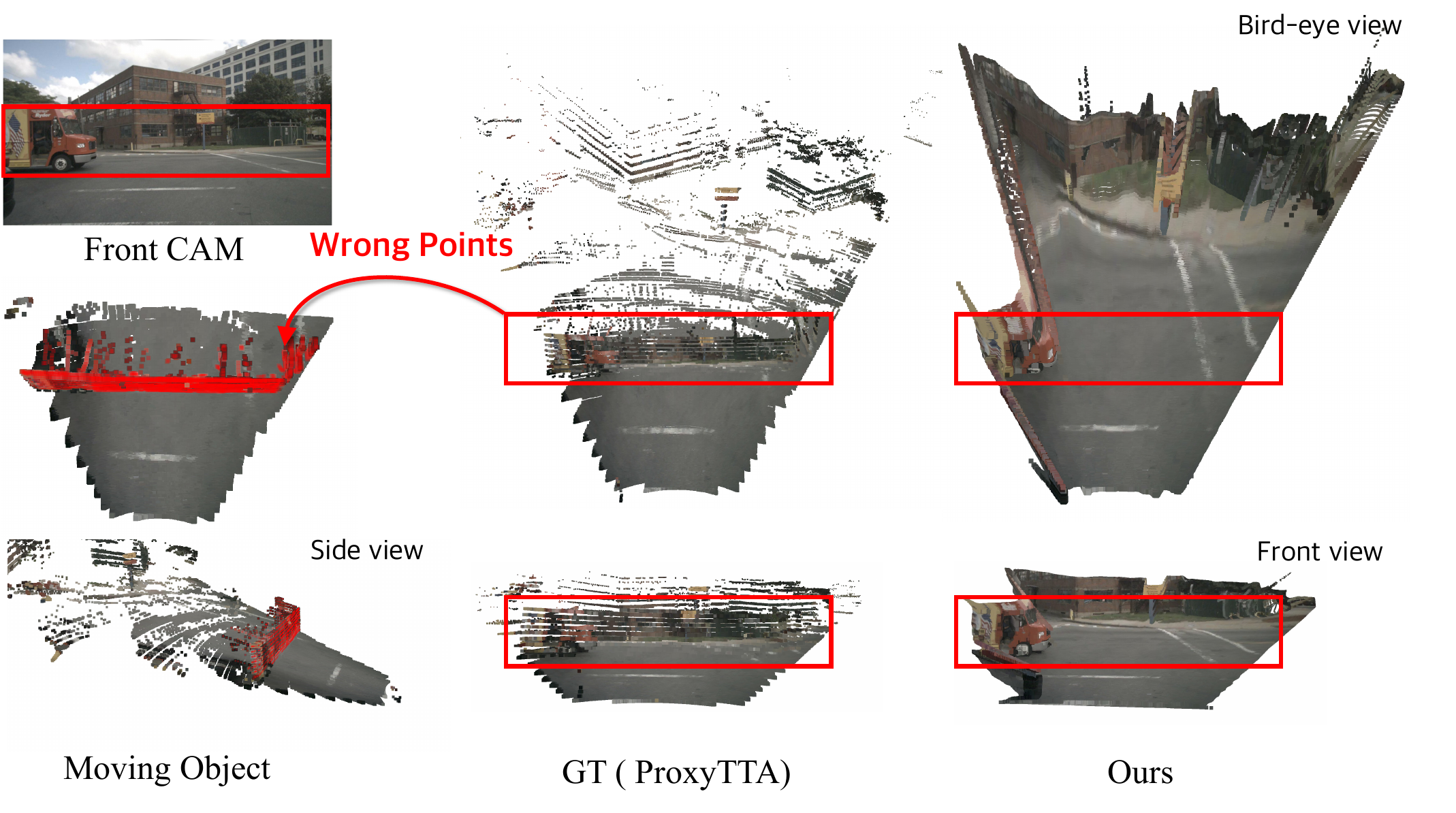}
   \caption{\textbf{Physically inaccurate ProxyTTA GT sample 1.} A moving truck is not detected by the off-the-shelf model, resulting in a high-error region.} 
   \label{fig:wrong_1}
\end{minipage}

\vspace{10mm} 

\begin{minipage}{\textwidth}
   \centering
   \includegraphics[width=0.9\linewidth]{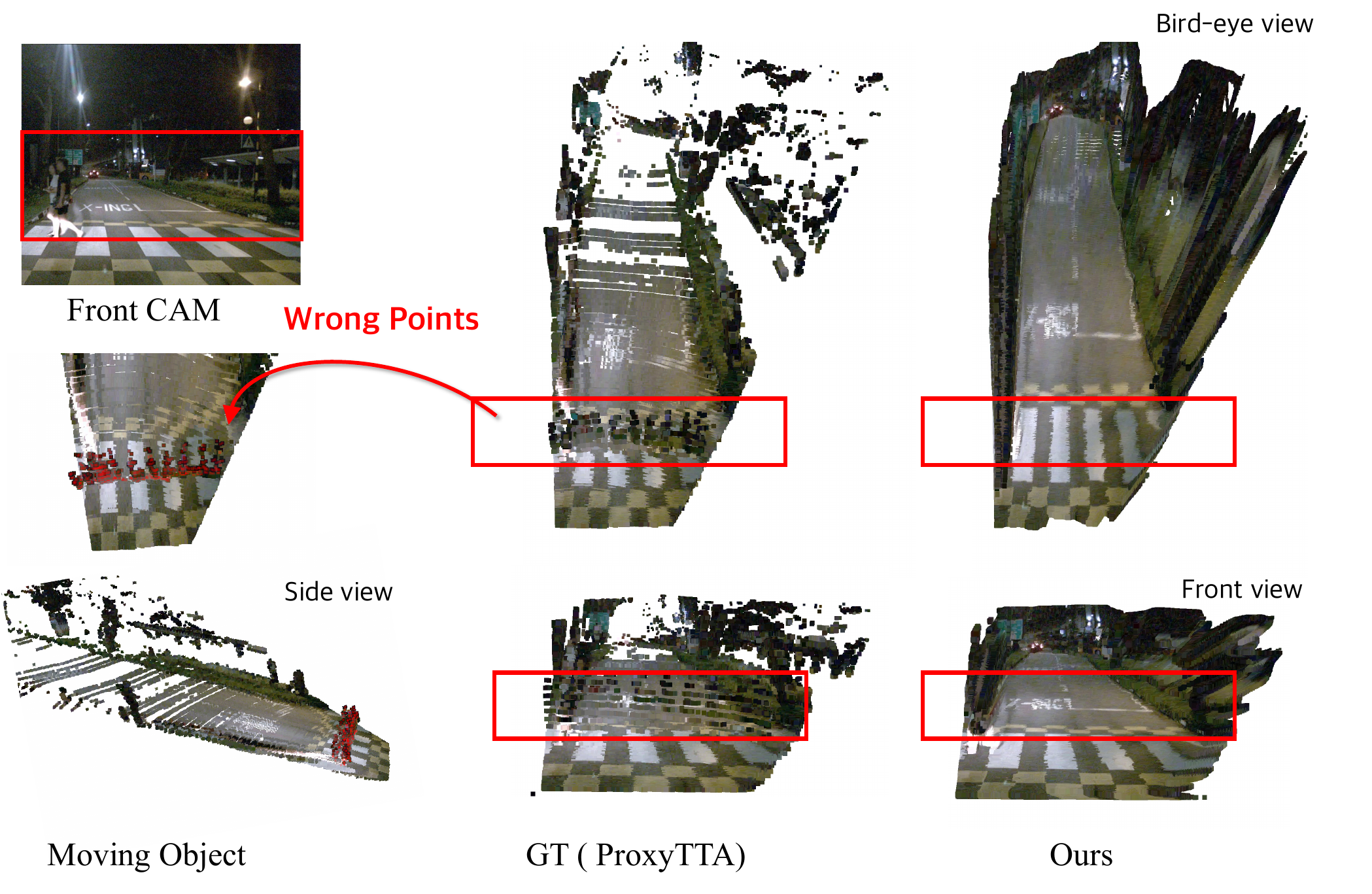}
   \caption{\textbf{Physically inaccurate ProxyTTA GT sample 2.} A walking human is not detected by the off-the-shelf model, resulting in a high-error region.} 
   \label{fig:wrong_2}
\end{minipage}
\end{figure*}
\subsection{Dataset Configurations}
For the domain generalization experiments, we use NYUv2~\cite{silberman2012nyu} and SceneNet~\cite{mccormac2017scenenet} as indoor datasets and nuScenes~\cite{caesar2020nuscene} and Waymo~\cite{sun2020waymo} as outdoor datasets.
We strictly follow the dataset configurations for the test-time scenario as suggested in ProxyTTA~\cite{park2024testtime}.
For indoor datasets, sparse depth maps are generated using a SLAM/VIO style with the Harris corner detector~\cite{harris1988combined}, based on dense depth maps acquired from RGB-D sensors like the Microsoft Kinect or simulation systems.
For outdoor datasets, sparse depth maps are acquired through long-range sensors, such as LiDAR.

\begingroup
\setlength{\tabcolsep}{11pt}
\begin{table}[t]
\centering
\renewcommand{\arraystretch}{1.4} 
    \resizebox{1.0\linewidth}{!}{
    \begin{tabular}{m{2.5cm} cc cc }
    \toprule
    \multirow{2}[2]{*}{Method} & \multicolumn{2}{c}{PCACC} & \multicolumn{2}{c}{ProxyTTA}\\
     \cmidrule(lr){2-3}  \cmidrule(lr){4-5} 
    & RMSE & MAE & RMSE & MAE \\
    \midrule 
     Pre-trained 
     & 3.998 & 1.967 & 6.630 & 3.064 \\ 
     BNAdapt
     & 1.801 & 0.828 & 6.391 & 2.306\\ 
     CoTTA
     & 2.668 & 1.222 & 6.099 & 2.676\\ 
     ProxyTTA
     & 1.755 & 0.799 & 5.509 & 2.062\\ 
    \midrule
    Ours
    & 1.516 & 0.561 & 5.876 & 2.499\\
    \bottomrule
    \end{tabular}
    }
\caption{\textbf{Quantitative results on the nuScenes depth completion benchmarks.} We evaluate our method on the nuScenes dataset using both PCACC~\cite{huang2022pcacc} and ProxyTTA~\cite{park2024testtime} ground truth datasets, employing the CostDCNet~\cite{kam2022costdcnet} model pre-trained on KITTI DC. Excluding ours, other test-time adaptation methods are adapted with CostDCNet. Our method shows favorable performance on the outdoor dataset and demonstrates domain generalizability, by evaluating on the more physically accurate benchmark.}
\label{tab:nu_table}
\vspace{-5pt}
\end{table}
\endgroup

\section{Additional Experiments}
In this section, we provide the additional experiments and analyses that complement our main paper.
First, we discuss why we use ground truth processing method from \citet{huang2022pcacc} rather than \citet{park2024testtime} for the nuScenes~\cite{caesar2020nuscene} dataset benchmark.
Second, we handle compatability of affine-invariant depth diffusion model for metric depth, discussed in the main paper, and analyze the potential issues of depth diffusion model's stochastic nature in deterministic dense prediction tasks such as depth estimation and completion. 
Lastly, we detailed describe our prior-based outlier filtering method and additional results demonstrating our method's effectiveness.

\begin{figure*}[!ht]
\centering
   \includegraphics[width=1.0\linewidth]{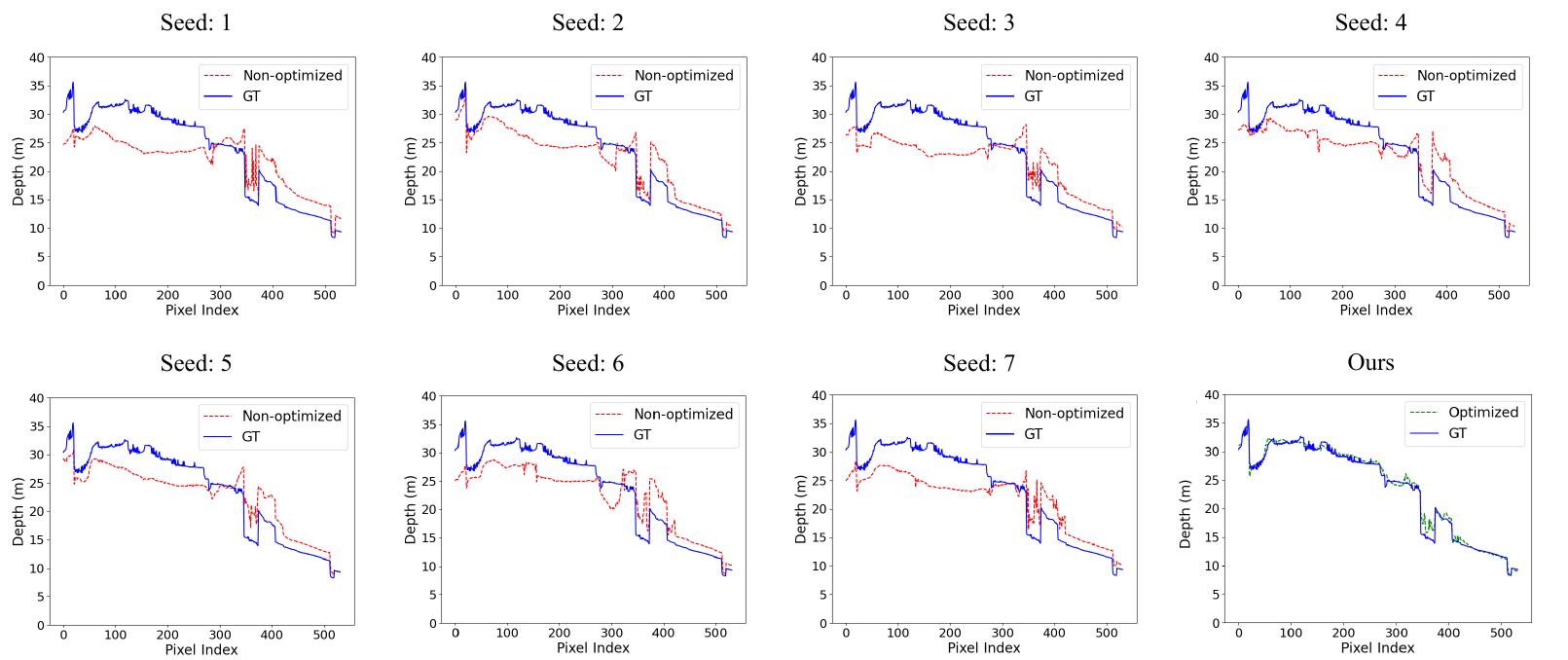}
   \caption{\textbf{Stochastic nature of depth diffusion model.}
   Due to the diffusion model’s stochastic process, the outcomes vary depending on the different seeds used.
   Our test-time alignment method effectively handles uncertainty derived from the stochastic process of the diffusion model.} 
\label{fig:stochastic}
\vspace{-2mm}
\end{figure*}
\begin{table}[!t]
\centering
\renewcommand{\arraystretch}{1.5} 
\large
    \resizebox{1.0\linewidth}{!}{
    \begin{tabular}{m{3cm} c c ccc}
    \toprule
    \multirow{2}{*}{Methods} & \multirow{2}{*}{Initial estimation} & \multicolumn{4}{c}{\makecell{Reconstruction \\ (timestamps)}} \\
    \cmidrule(lr){3-6}
    && 0  & 50 & 200 & 1000 \\ 
    \midrule
    KBNet & 1.126 & 1.198 & 1.214 &1.303 & 3.475   \\
    CompletionFormer  & 0.708 & 0.885 & 0.926 & 1.059 & 3.443\\
    \bottomrule
    \end{tabular}
    }
\caption{\textbf{{Capacity of depth diffusion prior representing metric depth.}} By reconstructing initial estimation from the pre-trained depth completion model using the affine-invariant depth estimation model, this affine-invariant depth prior has the potential to handle normalized metric depth space.}
\label{tab:reconstuction}
\vspace{-2mm}
\end{table}

\subsection{Sensitivity of the Ground Truth Processing of nuScenes Benchmark}
As mentioned in \Sref{sec:exp_domain} of the main paper, the ground truth dataset can vary depending on the accumulation method for LiDAR points and the moving object point removal method.
In Table 1 of the main paper, we report the generalization performance on the nuScenes ground truth data obtained 
by the method of ProxyTTA~\cite{park2024testtime}. This ground truth data is obtained by preprocessing the test split dataset of nuScenes, which involves accumulating subsequent frames and removing moving objects using off-the-shelf models.
However, we observe that the off-the-shelf models sometimes fail to detect and remove moving objects, leading to physically inaccurate ground truth depth. Figures~\ref{fig:wrong_1} and \ref{fig:wrong_2} demonstrate this failure case. In the nuScenes dataset, the 3D-lifted ground truth depth by ProxyTTA represents 3D points from moving trucks that are closer than distant walls as ground truth. Such errors can lead to depth discrepancies of up to 10-20 meters in some samples, which likely contribute to the high RMSE values of 5-6 meters reported in \Tref{tab:nu_table} of the main paper.

\para{Evaluation on physically accurate benchmark}
\citet{huang2022pcacc} provide a nuScenes semi-dense depth (\ie, ground truth) dataset based on the validation split by accumulating frames and removing moving objects using manually annotated bounding boxes. This dataset, which relies on manual annotation, is free from the failures of off-the-shelf models and is physically accurate.
Using the nuScenes ground truth provided by \citet{huang2022pcacc} (PCACC), we assess the domain generalization performance of our method and previous test-time adaptation methods~\cite{wang2021tent,wang2022continual, park2024testtime}. 

In this experiment, the competing test-time adaptation methods use pre-trained depth completion model CostDCNet~\cite{kam2022costdcnet} trained in KITTI DC~\cite{uhrig2017sparsity} for adaptation. To independently evaluate the impact of ground truth acquisition methods, we also report the performance on the ground truth of ProxyTTA. 
Table~\ref{tab:nu_table} summarizes the results. When using the PCACC ground truth instead of that of ProxyTTA, we observe the trend of overall metric improvement across all methods, likely due to the higher physical accuracy of the ground truth. Additionally, when comparing with other competing test-time adaptation methods, our method achieves the best performance.

\begin{figure*}[t]
\centering
   \includegraphics[width=0.9\linewidth]{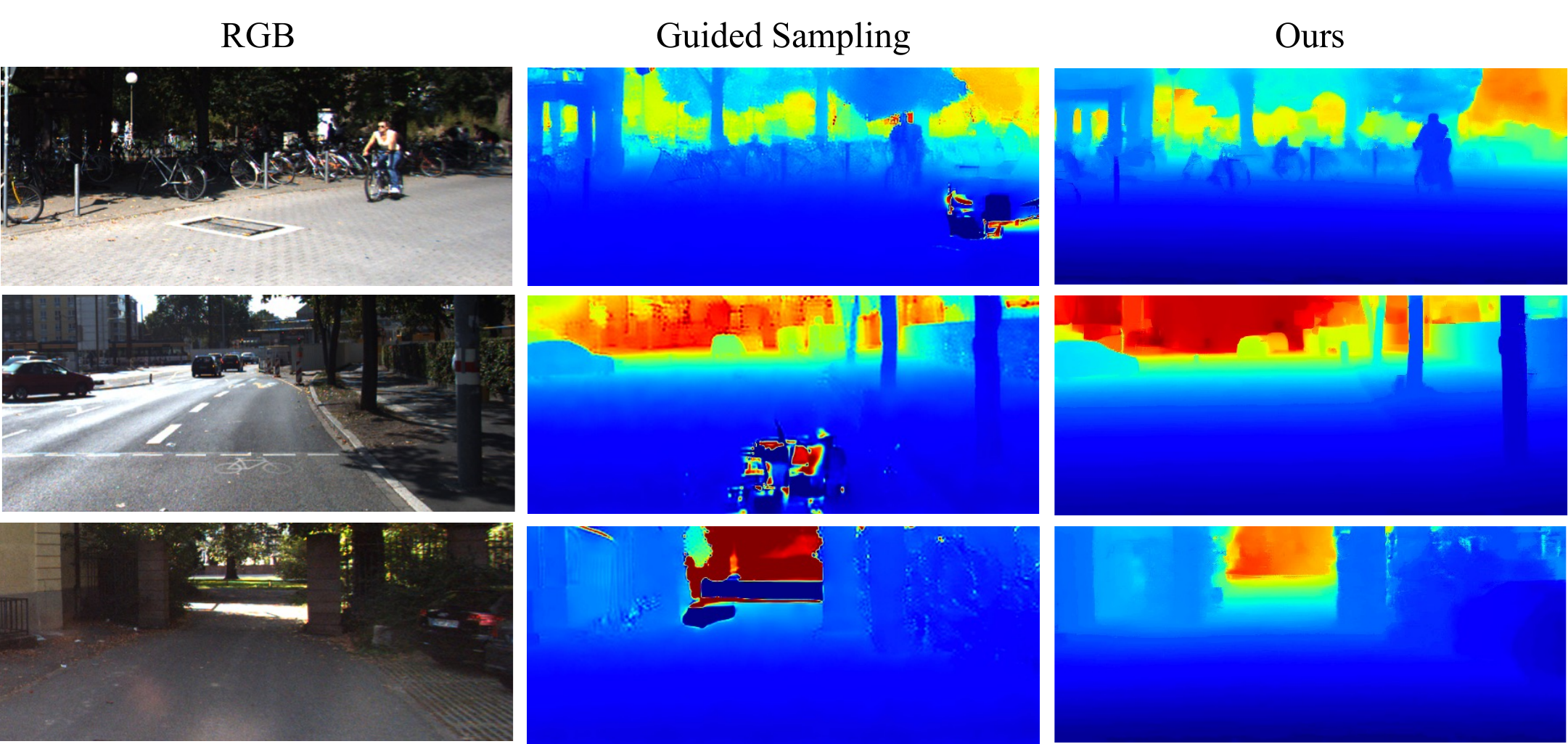}
   \caption{\textbf{Comparison of guided sampling and our test-time alignment methods.} When using guided sampling~\cite{chung2023dps} with sparse measurements, the completed depth map often becomes corrupted due to the stochastic nature of the diffusion model. In contrast, our test-time alignment method directs the stochastic process towards a desirable solution that aligns with the sparse measurements.} 
\label{fig:unintended}
\end{figure*}

\subsection{Analysis of Test-Time Alignment Method}
\para{Compatibility of Depth Diffusion Prior with Metric Depths 
}
As mentioned in \Sref{sec:opt_sampling} of the main paper, we investigate whether the normalized metric depth space can be represented by 
the depth diffusion models~\cite{ke2023repurposing, gui2024depthfm}, which are trained only on synthetic data.
These can be empirically verified
by checking the consistency of the normalized metric depth map with the depth map reconstructed through the reverse sampling of the diffusion model.
To verify this, we obtain the normalized metric depth maps using two existing depth completion models, \ie, KBNet and CompletionFormer~\cite{wong2021unsupervised,zhang2023completionformer}. Then, we reconstruct the metric depth maps after applying different noise levels and reverse sampling, so that we can see whether those metric depths can be re-represented by the depth diffusion model~\cite{ke2023repurposing}, \ie, lie in our prior space. \Tref{tab:reconstuction} shows the RMSE between the metric depth map and ground truth, as well as between the reconstructed depth map and ground truth. 
For simplicity, we denote the noise level by the DDIM sampler's timestamp, \ie, larger timestamps correspond to higher noise levels. 
The reconstructed depth map shows similar performance to the metric depth map up to timestamp 200 while achieving significantly better performance than the starting from random noise, \ie, timestamp 1,000.
This suggests that the affine-invariant depth prior we used is sufficient to well represent normalized metric depth.

\para{Potential problem of stochastic process}
As discussed in \Sref{sec:opt_sampling} of the main paper, we highlight the stochasticity introduced by the diffusion model's stochastic process and the associated potential risk of falling into unintended solutions.
In this section, we experimentally show this stochastic behavior and its potential to lead to undesirable results.

Since the diffusion model starts from random noise, its outputs vary depending on the initial noise, leading to different results with each run. 
This stochasticity can produce unintended outcomes in cases where a deterministic solution exists, such as depth estimation and completion. 
Figure~\ref{fig:stochastic} shows how altering only the initial noise can result in different outputs for the same sample.
Furthermore, by simply performing guided sampling~\cite{chung2023dps}, we confirm that the potential risks discussed in \Sref{sec:opt_sampling} can lead to undesirable solutions.
Furthermore, as shown in \Fref{fig:unintended}, we observe that guided sampling leads to undesirable solutions where the depth map becomes corrupted, as discussed in \Sref{sec:opt_sampling} of the main paper.
These experimental results not only support the potential risks mentioned in the main paper but also highlight the necessity of our hard constraints and correction steps.

\begin{figure*}[t]
\centering
   \includegraphics[width=1.0\linewidth]{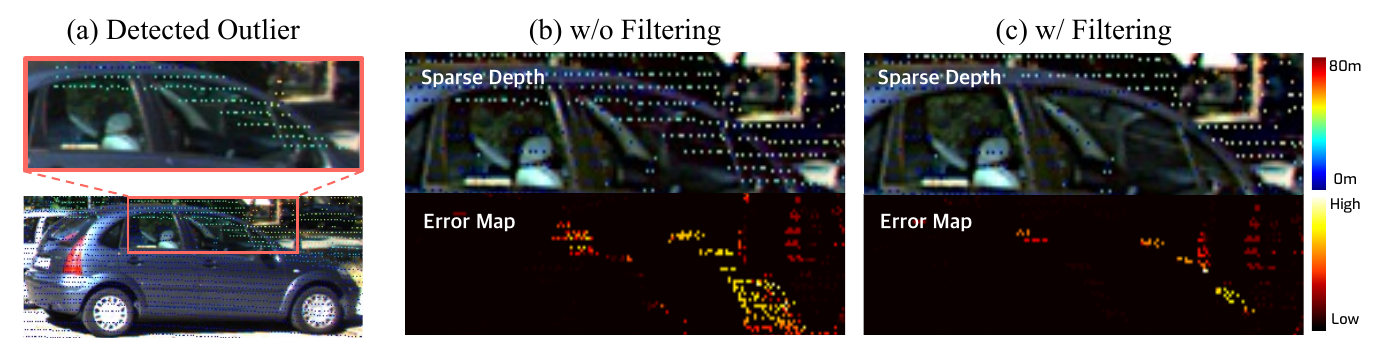}
   \caption{\textbf{Effectiveness of prior-based outlier filtering.}
    Our outlier filtering algorithm effectively detects see-through points on car windows in sparse depth measurements.
    Aligning with filtered sparse depth and visualizing the absolute error map against the ground truth shows the elimination of high-error regions caused by outliers,  specifically see-through points on car windows.
   } 
\label{fig:outlier}
\end{figure*}
\subsection{Analysis of Outlier Filtering Method}
We evaluate our outlier filtering algorithm on the KITTI DC validation set~\cite{uhrig2017sparsity}
by computing the Area Under the Sparsification Curve (AUC), a standard metric for assessing the reliability of outlier detection confidence in LiDAR depth maps~\cite{conti22confidence, ilg2018uncertainty}, as shown in \Tref{tab:noise_filter}. 
For evaluation, we apply the outlier filtering algorithm to each component: the sparse depth map from a synchronized single frame and the accumulated semi-dense depth map before processing the accumulation.
Each filtered depth map is evaluated against the sparse and semi-dense ground truth, derived from the manually processed semi-dense depth map in KITTI DC.
For measuring AUC, pixels with both single-frame sparse depth and accumulated semi-dense depth are sorted by confidence and removed incrementally. 
We define confidence as $|\hat{\mathbf{y}}_i-\mathbf{y_i}|$, normalized to a 0-1 range in each segment.
In \Tref{tab:noise_filter}, RMSE is calculated on the remaining pixels to draw a curve, with AUC (lower values indicate better performance) measuring outlier removal effectiveness.
Our prior-based outlier filtering algorithm outperforms the commonly used method that removes distant points as outliers using a shifting window~\cite{lopezrodriguez2020project, wong2021unsupervised} for both sparse and semi-dense depth.
Additionally, our approach outperforms in sparse depth outlier filtering and performs favorably on semi-dense depth maps compared to recent methods using learning-based confidence estimation for outlier removal.
Unlike these methods, which rely on in-domain training and may not be applicable to other datasets, our approach is more adaptable to any domain, \ie, zero-shot.
Figure \ref{fig:outlier} illustrates the importance of outlier filtering when using sparse depth supervision and demonstrates how effectively our prior-based outlier filtering detects these outliers.
\definecolor{tabfirst}{rgb}{1, 0.7, 0.7} 
\definecolor{tabsecond}{rgb}{1, 0.85, 0.7} 
\definecolor{tabthird}{rgb}{1, 1, 0.7} 

\begingroup
\begin{table}[!t]
\centering
\renewcommand{\arraystretch}{1.1} 
    \resizebox{1.0\linewidth}{!}{
    \begin{tabular}{m{2.6cm} ccc}
    \toprule
    \multirow{2}{*}{Filtering Method} & \multirow{2}{*}{In-domain} & {Sparse} & {Semi-dense}\\
    \cmidrule(lr){3-3}  \cmidrule(lr){4-4}
    & & AUC~($\downarrow$) & AUC~($\downarrow$)\\
    \midrule
    None & \xmark & 1.3541 & 2.5441  \\
    Window Filter & \xmark & \cellcolor{tabsecond}0.3480 & 0.9629  \\ 
    Ours & \xmark & \cellcolor{tabfirst}{0.2959} & \cellcolor{tabsecond}0.5103  \\ 
    \midrule
    Lidar Confidence & \cmark & 1.0521  & \cellcolor{tabfirst}{0.2117} \\
    \bottomrule
    \end{tabular}
    }
\caption{\textbf{Quantitative evaluation of outlier filtering.}
Our prior-based outlier filtering method demonstrates favorable
performance compared to existing methods.
In this table, ``None" denotes a synchronized sparse
and accumulated depth map
without any postprocessing.
Note that the ``Lidar Confidence'' method is a learning-based method trained on the same domain dataset with the evaluation dataset.
}
\label{tab:noise_filter}
\vspace{-4mm}
\end{table}

\begin{figure*}[t]
\centering
\begin{minipage}{\textwidth}
   \centering
   \includegraphics[width=0.85\linewidth]{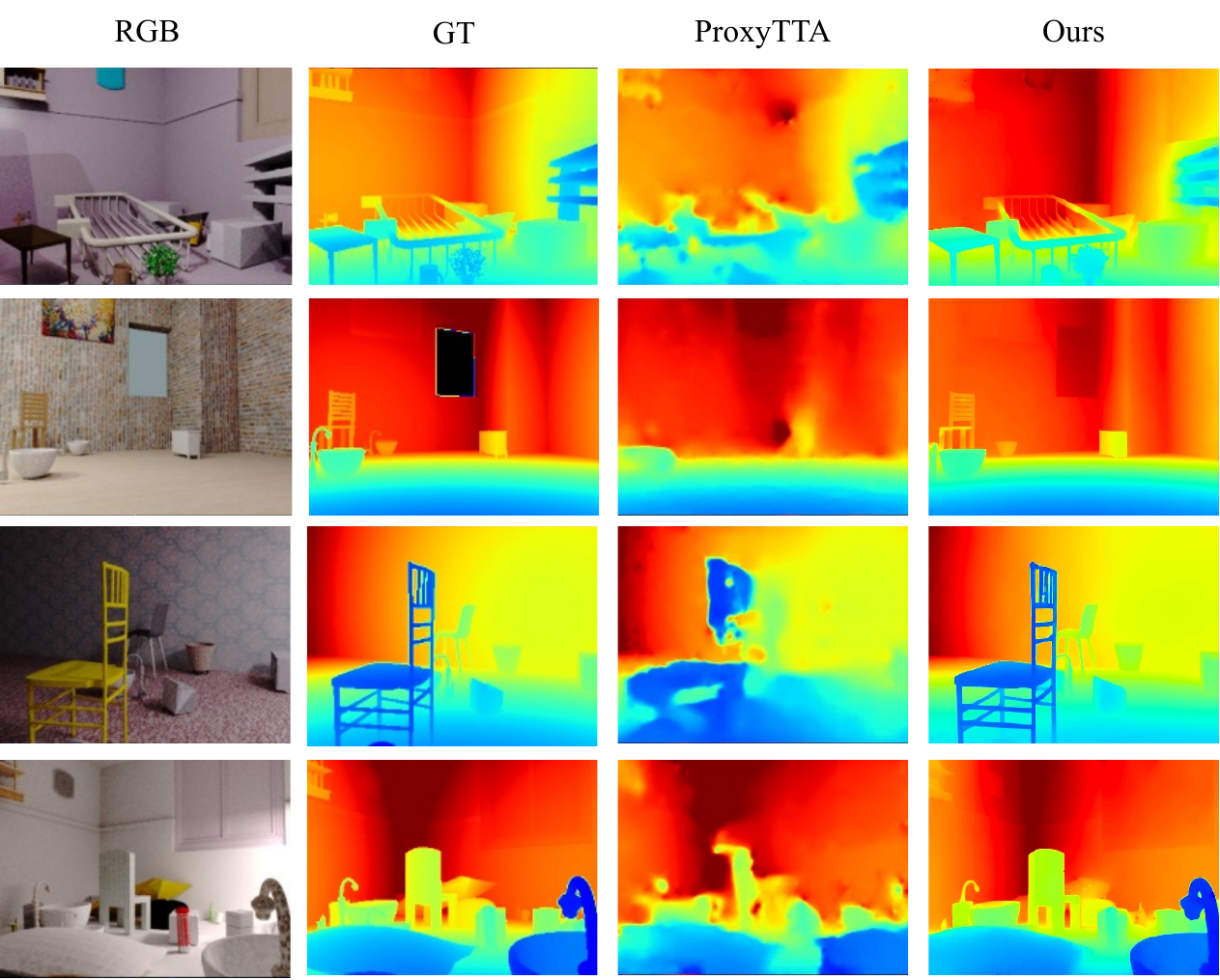}
   \caption{\textbf{Qualitative results on SceneNet.} Our zero-shot depth completion method outperforms the state-of-the-art test-time adaptation method by capturing the scene's structure effectively.} 
   \label{fig:qual_scenenet}
\end{minipage}

\vspace{10mm} 

\begin{minipage}{\textwidth}
   \centering
   \includegraphics[width=0.9\linewidth]{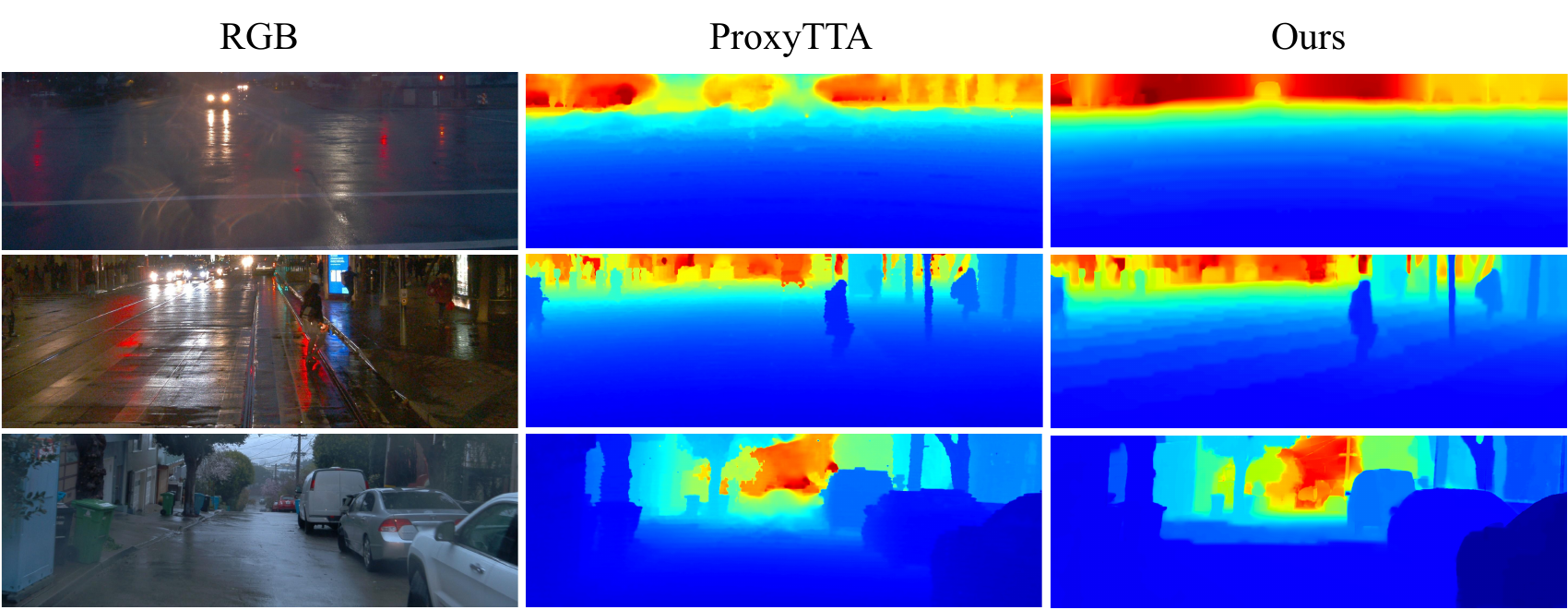}
   \caption{\textbf{Qualitative results on Waymo.} Our zero-shot depth completion method demonstrates robust performance even in extreme environments, such as rain or nighttime conditions.} 
   \label{fig:qual_waymo}
\end{minipage}
\end{figure*}

\section{Additional Qualitative Results}
In this section, we provide additional qualitative results corresponding to the experiments discussed in each subsection of the main paper.

\para{Domain generalization}
We provide additional qualitative results for dataset not covered in the main paper, such as SceneNet~\cite{mccormac2017scenenet} and Waymo~\cite{sun2020waymo}, in \Fref{fig:qual_scenenet} and \ref{fig:qual_waymo}.
Most existing pre-trained depth completion models tend to fail when faced with the difficult conditions typically encountered in real-world environments.
We also demonstrate the robust performance of our prior-based method in extreme environments, such as rain or nighttime, as shown in Fig.\ref{fig:qual_scenenet} and \ref{fig:qual_waymo}.
Additional results for these scenes will also be provided in the supplementary video.

\begin{figure*}[t]
\centering
   \includegraphics[width=1.0\linewidth]{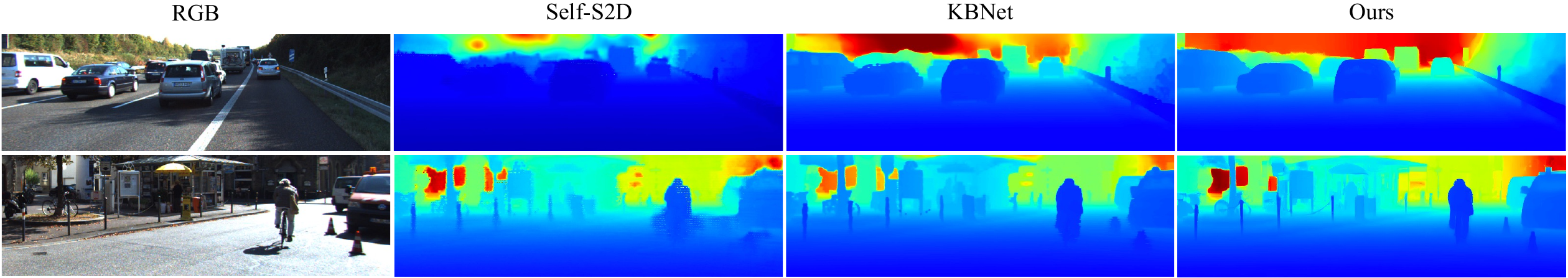}
   \caption{\textbf{Qualitative comparison on KITTI DC validation set.}
    Compared to the unsupervised methods~\cite{ma2018self, wong2021unsupervised} with comparable quantitative performance, our prior-based approach better preserves the scene structure and details.
   } 
\label{fig:unsup_out}
\end{figure*}

\para{Comparison with unsupervised methods on KITTI}
In \Fref{fig:unsup_out}, we provide a qualitative comparison on the KITTI DC dataset, an outdoor dataset not included in the main paper. Despite using only a monocular RGB view and sparse depth, unlike previous unsupervised methods~\cite{ma2018self, wong2021unsupervised}, we also complete a well-structured depth map


\end{document}